\documentclass[letterpaper, 10 pt, conference, twoside]{ieeeconf}  

\IEEEoverridecommandlockouts                              

\usepackage{hyperref} 
\hypersetup{
  colorlinks,
  citecolor=black,
  filecolor=black,
  linkcolor=black,
  urlcolor=black,
  pdfauthor={},
  pdfsubject={},
  pdftitle={}
}

\usepackage{graphicx}
\usepackage{setspace}
\usepackage{epstopdf}

\usepackage{wrapfig}
\usepackage{caption}
\usepackage{subcaption}
\usepackage{amssymb,amsmath}
\usepackage[export]{adjustbox}
\usepackage{tikz}
\usepackage{fancyhdr}
\usepackage{balance}

\usepackage{color}
\usepackage{url}

\usepackage{amsmath}
\usepackage{algorithm}
\usepackage{etoolbox}\AtBeginEnvironment{algorithmic}{\scriptsize}
\usepackage[noend]{algpseudocode}

\usepackage{booktabs}

\usepackage{listings}

\usepackage{numprint}

\usepackage{breqn}

\usepackage[utf8]{inputenc}

\usepackage{siunitx}
\makeatletter
\def\lst@makecaption{%
  \def\@captype{table}%
  \@makecaption
}
\makeatother

\makeatletter
\def\BState{\State\hskip-\ALG@thistlm}
\makeatother

\newcommand{\vect}[1]{\ensuremath{\mathbf{#1}}}

\newcommand{\module}[1]{\textit{#1}}

\newcommand{\m}[1]{\ensuremath{\mathbf{#1}}}
\newcommand{\vct}[1]{\ensuremath{\mathbf{#1}}}

\newcommand\numberthis{\addtocounter{equation}{1}\tag{\theequation}}

\newcommand{\figvspace}{\vspace{-1.0em}}

\newcommand{\rn}{\ensuremath{\mathbb{R}}}

\newcommand{\reffig}[1]{Fig.~\ref{#1}}

\newcommand{\reftab}[1]{Table~\ref{#1}}

\newcommand{\map}{\ensuremath{\mathbf{M}}}
\newcommand{\tf}{\ensuremath{\mathbf{T}}}
\newcommand{\point}{\ensuremath{\mathbf{p}}}
\newcommand{\pr}{\ensuremath{\mathbf{q}}}
\newcommand{\ps}{\ensuremath{\mathbf{p}}}
\newcommand{\lscan}{\ensuremath{\mathbf{Z}}}

\newcommand{\uavsub}[1]{\ensuremath{#1}_{uav}}

\newcommand{\refsection}[1]{Section~\ref{#1}}

\newcommand{\deltapos}{\ensuremath{\Delta{\vect{x}}}}
\newcommand{\posref}{\ensuremath{\vect{x}_{t_1}}}
\newcommand{\posact}{\ensuremath{\vect{x}_{t_2}}}
\newcommand{\inr}{\ensuremath{\in \rn}}

\usepackage{tikz}
\usepackage{pgfplots}
\usetikzlibrary{backgrounds,arrows,automata,shapes,positioning,calc,through}
\pgfdeclarelayer{background}
\pgfdeclarelayer{foreground}
\pgfsetlayers{background,main,foreground}

\tikzset{
  state/.style={
    rectangle,
    draw=black, very thick,
    minimum height=1.0em,
    text centered,
  },
  final_state/.style={
    rectangle,
    rounded corners,
    draw=black, very thick,
    minimum height=2em,
    text centered,
  },
  initial_state/.style={
    rectangle,
    double=white,
    double distance=1pt,
    inner sep=2pt,
    draw=black, very thick,
    minimum height=2em,
    text centered,
  },
  point/.style={
    circle,
    inner sep=0pt,
    minimum size=3pt,
    fill=red
  },
  adder/.style={
    circle,
    inner sep=2pt,
    minimum size=0.3in,
    draw=black, very thick,
    text centered
  }
}

\usepackage{tikz}
\usepackage{pgfplots}
\usetikzlibrary{backgrounds,arrows,automata,shapes,positioning,calc,through}
\pgfdeclarelayer{background}
\pgfdeclarelayer{foreground}
\pgfsetlayers{background,main,foreground}

\tikzset{
  state/.style={
    rectangle,
    draw=black, very thick,
    minimum height=1.0em,
    text centered,
  },
  final_state/.style={
    rectangle,
    rounded corners,
    draw=black, very thick,
    minimum height=2em,
    text centered,
  },
  initial_state/.style={
    rectangle,
    double=white,
    double distance=1pt,
    inner sep=2pt,
    draw=black, very thick,
    minimum height=2em,
    text centered,
  },
  point/.style={
    circle,
    inner sep=0pt,
    minimum size=3pt,
    fill=red
  },
  adder/.style={
    circle,
    inner sep=2pt,
    minimum size=0.3in,
    draw=black, very thick,
    text centered
  }
}

\title{\LARGE \bf LIDAR-based Stabilization, Navigation and Localization for UAVs Operating in Dark Indoor Environments}

\author{Mat\v ej Petrl\' ik$^{*}$, Tom\' a\v s Krajn\' ik$^{*}$ and Martin Saska$^{*}$%
  \thanks{The presented work has been supported by the Czech Science Foundation (GAČR) under research project No. 20-29531S, by CTU grant no SGS20/174/OHK3/3T/13, and by OP VVV funded project CZ.02.1.01/0.0/0.0/16 019/0000765 "Research Center for Informatics". 
  This project has also received funding from the EU H2020 project AERIAL CORE, No. 871479.}
 \thanks{$^{*}$Authors are with the Faculty of Electrical Engineering,
    Czech Technical University in Prague, Technicka 2, Prague 6,
  Email: {\tt\small \{matej.petrlik, tomas.krajnik, martin.saska\} @fel.cvut.cz}.}%
}

\begin{document}

\newcommand{\PREPRINTYEAR}{2021}
\newcommand{\PREPRINTPUBLISHER}{IEEE}

\onecolumn
\pagenumbering{gobble}
{
  \topskip0pt
  \vspace*{\fill}
  \centering
  \LARGE{%
    \copyright{} \PREPRINTYEAR~\PREPRINTPUBLISHER\\\vspace{1cm}
	Personal use of this material is permitted.
	Permission from \PREPRINTPUBLISHER~must be obtained for all other uses, in any current or future media, including reprinting or republishing this material for advertising or promotional purposes, creating new collective works, for resale or redistribution to servers or lists, or reuse of any copyrighted component of this work in other works.}
	\vspace*{\fill}
}

\twocolumn 
\pagenumbering{arabic}


\fancypagestyle{empty}{
\setlength{\headsep}{10pt}
  \fancyhead{}
  \fancyfoot{}
\renewcommand{\headrulewidth}{0pt}
\fancyhead[LO,RE]{\footnotesize \copyright{} \PREPRINTPUBLISHER, \PREPRINTYEAR. Accepted to ICUAS 2021. DOI: \href{https://doi.org/10.1109/ICUAS51884.2021.9476837}{10.1109/ICUAS51884.2021.9476837}}
\fancyhead[LE,RO]{\footnotesize \thepage}
}

\maketitle
\global\csname @topnum\endcsname 0
\global\csname @botnum\endcsname 0
\thispagestyle{empty}
\pagestyle{empty}



\begin{abstract}

  Autonomous operation of UAVs in a closed environment requires precise and reliable pose estimate that can stabilize the UAV without using external localization systems such as GNSS.
  In this work, we are concerned with estimating the pose from laser scans generated by an inexpensive and lightweight LIDAR.
  We propose a localization system for lightweight (under \SI{200}{\gram}) LIDAR sensors with high reliability in arbitrary environments, where other methods fail.
  The general nature of the proposed method allows deployment in wide array of applications. 
  Moreover, seamless transitioning between different kinds of environments is possible.

  The advantage of LIDAR localization is that it is robust to poor illumination, which is often challenging for camera-based solutions in dark indoor environments and in the case of the transition between indoor and outdoor environment.
  Our approach allows executing tasks in poorly-illuminated indoor locations such as historic buildings and warehouses, as well as in the tight outdoor environment, such as forest, where vision-based approaches fail due to large contrast of the scene, and where large well-equipped UAVs cannot be deployed due to the constrained space.

\end{abstract}

\section*{Multimedia material}
\begin{center}
  A video attachment to this work is available at: {\small \url{http://mrs.felk.cvut.cz/icuas2021lidar}}
\end{center}


\section{Introduction}


\begin{figure}
  \centering
      \begin{subfigure}{0.49\linewidth}
        \centering
        \adjincludegraphics[width=1.0\linewidth, trim={{0.25\width} {0.257\height} {0.25\width} {0.15\height}}, clip=true]{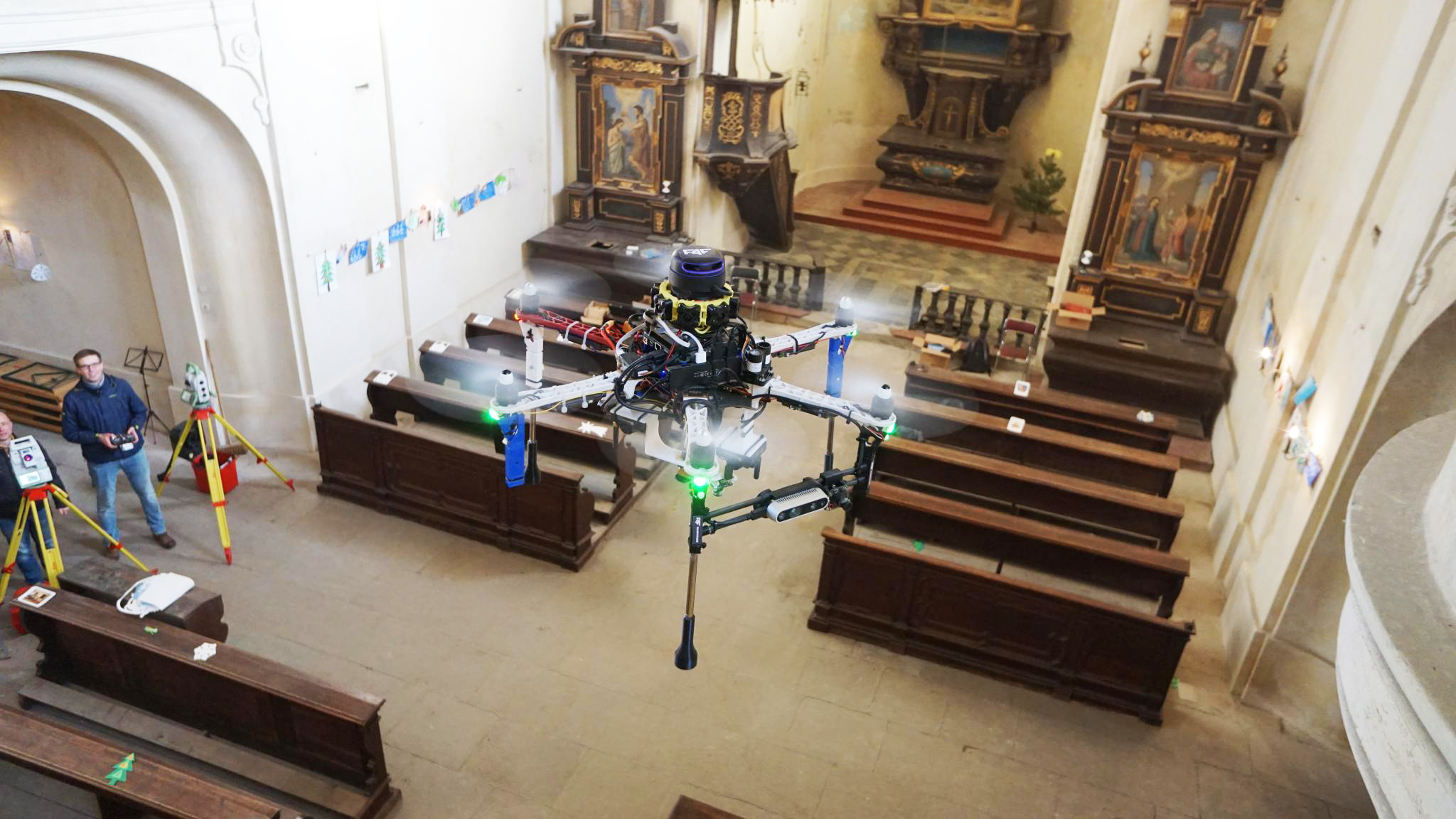}
      \end{subfigure}
      \begin{subfigure}{0.49\linewidth}
        \centering
        \adjincludegraphics[width=1.0\linewidth, trim={{0.0\width} {0.05\height} {0.1\width} {0.05\height}}, clip=true]{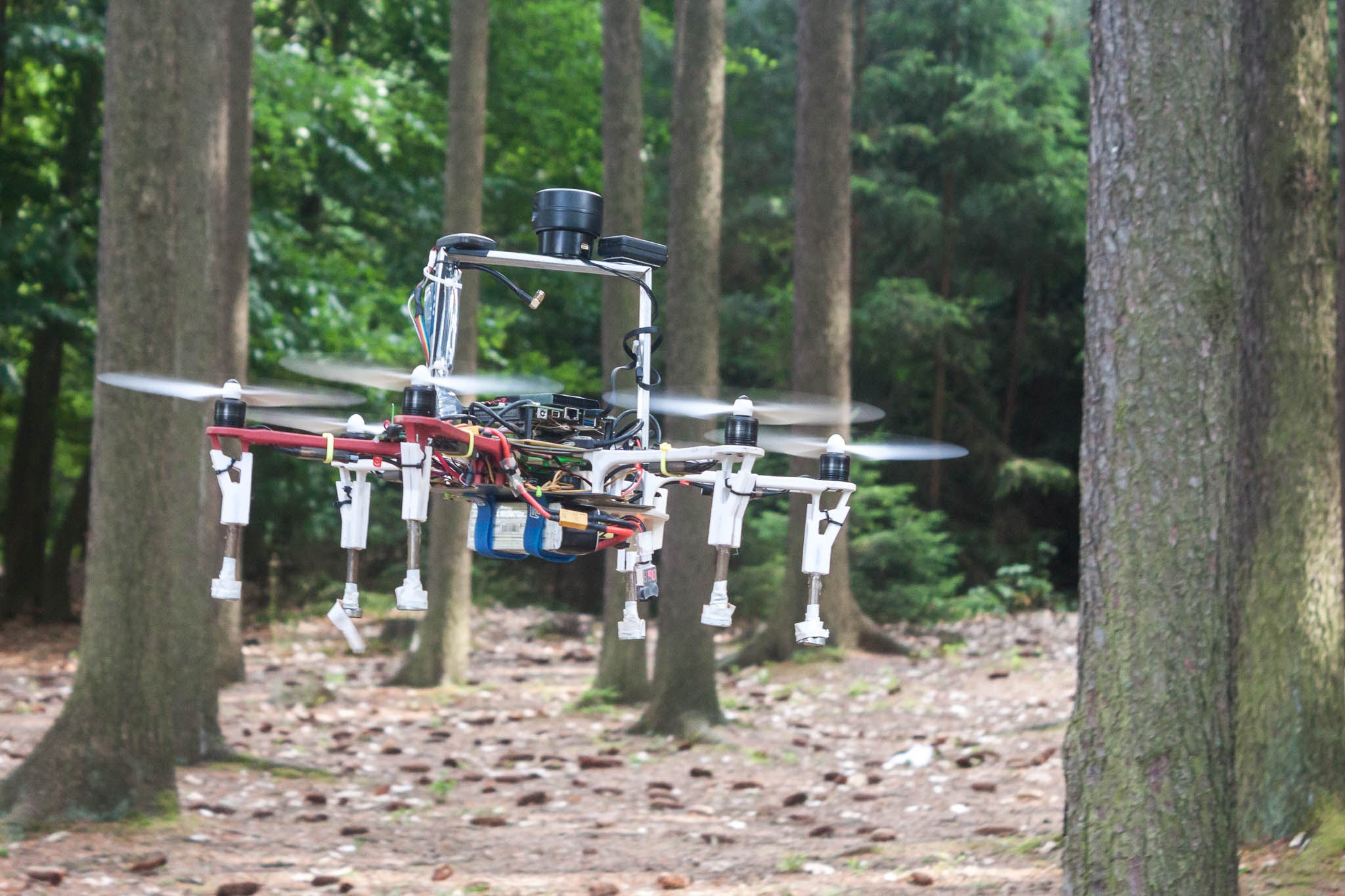}
      \end{subfigure}
  \caption{
    The left image shows our UAV deployed to document the interiors of the Chlumin church for the historians.
    In the right image, the UAV is navigating through a forest.
    \label{fig:intro}
  }
  \figvspace
\end{figure}

Estimating the pose of a UAV is a challenging problem due to limited weight capacity, which implies restricted processing power and lightweight imperfect sensors.
The proposed localization system is designed with the limitations of micro-scale UAVs in mind.
The method runs online onboard the UAV to provide real-time estimates of the UAV pose which are fed back to the controller to stabilize the UAV.
A full 3D space pose estimate can be obtained even using a cheap, small and lightweight 2D LIDAR thanks to the proposed robust data fusion approach.
Compared to state-of-the-art visual odometry approaches, our method is primarily targeted for use in compact environments with high density of obstacles such as for example historic buildings and forests, where cameras cannot be used due to the poor lighting conditions.
The ability to stabilize a UAV is proven by experimental verification with real UAV platform in real-world conditions.

The control algorithms of unmanned aerial vehicles (UAVs) are precise enough for deployment in tight, enclosed environments with obstacles, which enables inspection, documentation and monitoring of indoor areas of warehouses, historic buildings, churches, or temples.
In particular, the non-linear \textit{SE(3)} state feedback controller based on the work in \cite{lee2010geometric} and \cite{mellinger2010minimum} is widely adopted by robotic teams working with UAVs.
The authors of \cite{harik2016warehouse} propose to use a heterogeneous team consisting of a UGV and UAV to automatize the inventory process in a warehouse.
The UGV carries the UAV to a rack of goods, where the UAV takes off and scans the codes of the goods in the rack and then lands on the UGV again which moves to a next rack.
In \cite{saska_etfa17}, a team of autonomous UAVs is used to inspect the Saint Nicholas church in Prague and Virgin Mary church in Sternberk, Czech Republic by applying illumination techniques consisting of a camera and multiple light sources carried by a self-stabilized team of UAVs.
The methods of predictive control are used to optimize the angles and ranges of light sources while taking the motion and formation constraints into account.

To reliably navigate through an environment, a precise state estimate is required to follow the desired trajectory successfully while avoiding surrounding obstacles.
This work is concerned with localization of the UAV, i.e., obtaining the state estimate from available onboard sensors.
The most often used localization technique is based on Global Navigation Satellite System (GNSS) technologies.
These techniques require an open environment with satellite visibility and even then provide pose estimate with a standard deviation of several meters.

In GNSS-denied locations (indoor environment, close to large buildings, forests, mines, caves) the UAVs are often localized using an image stream from one or more onboard cameras.
Especially visual simultaneous localization and mapping (SLAM) \cite{cadena2016slam}, and visual inertial odometry (VIO) \cite{delmerico2018vio} methods have received a lot of attention thanks to the pose error reduction during loop closure and robustness to lost feature tracking respectively.
However, cameras require favorable lighting conditions to function optimally \cite{alismail2017lowlight}, due to the limited dynamic range of the sensor.
The worst scenarios for camera-based localization are:
\begin{enumerate}
  \item Dark areas with insufficient illumination, typically churches, ruins, mines, tunnels, or any tasks performed in the night cause underexposure of the whole image.
  \item Source of light pointing directly into the camera, which happens during outdoor missions when horizontally mounted cameras are pointed into the sun, but can also happen in buildings with windows or bright lighting sources.
  \item  High contrast scenes where part of the image is overexposed and part underexposed, which often occurs in forests - one of our target applications.
\end{enumerate}

The problem of insufficient illumination can be overcome by using structured light or Time-of-Flight cameras.
These cameras allow localization in total darkness at the cost of reduced resolution and range, with most of them becoming unusable in outdoor applications due to sunlight \cite{horaud2016tof}.

We propose a solution that provides the pose estimate by aligning laser scans obtained from a small 2D LIDAR.
When operating in a tight space, the size of the UAV is limited, so only low-weight payload can be carried.
It implies that heavy sensors such as the Velodyne PUCK that provide 3D scans of the surroundings cannot be used.
Lightweight LIDARs (Scanse Sweep \footnote{Scanse Sweep v1 LIDAR, \url{https://s3.amazonaws.com/scanse/Sweep_user_manual.pdf}}, SLAMTEC RPLIDAR \footnote{SLAMTEC RPLIDAR \url{https://www.generationrobots.com/media/LD310_SLAMTEC_rplidar_datasheet_A3M1_v1.0_en.pdf}}) are more suitable for this task, since they can be carried by small UAVs, and are less expensive, which makes them widely used.
However, the small form factor also means a lower number of points per scan, lower scanning frequency, lower range, noise with higher variance and more false detections.
Most importantly, the sensor provides 2D scans, while the UAV is a 6 DOF robot moving through a 3D environment.
As a result, the raw 2D laser scan without compensating the UAV tilt does not represent the surroundings of the UAV accurately when placed in the 3D space, and the full position estimate cannot be obtained without measurements from other sensors.
We address these issues in the proposed system by modifying and combining multiple state-of-the-art algorithms so that they work well with the available data.



\section{Related work}

In \cite{opromolla2016lidar} the authors present autonomous localization and mapping in GNSS denied environments for UAVs.
They use IMU and LIDAR measurements to estimate the position by matching consecutive scans while simultaneously building a map.
The current laser scan is matched to the previous one by a customized ICP algorithm to obtain the current position of the UAV.
For building the map, Principal Component Analysis (PCA) was used.
The method uses spectral decomposition of the covariance matrix of clusters in the laser scan to fit line segments to them.
Although the proposed method has a similar motivation as the system proposed in this paper, it has a few shortcomings compared to our approach.
The sequential localization that is prone to drift is not corrected by the gradually built map. 
The map is not scalable to arbitrary environments as it assumes that the geometry of the environment consists of only line features.
Moreover, the experimental verification is done by post-processing recorded datasets, which does not guarantee real-time performance.

In contrast to \cite{opromolla2016lidar}, we present a clear demonstration of applicability to our platform in \refsection{sec:experiments}.
Our solution in addition to the tasks solved in \cite{opromolla2016lidar} contains the fusion of position obtained from localization in a global map, velocity from the displacement of sequential scans and attitude measurements from UAV gyroscopes.

A comprehensive control, localization, navigation, and mapping solution for indoor UAV is introduced in \cite{wang2013comprehensive}.
The controller accepts state estimates from a linear Kalman filter which is fusing IMU accelerations with velocities obtained by computing the optical flow of a downward facing camera.
The position estimate can, however, drift over time, since no measurement of absolute position is fused in the Kalman filter.
The authors suppress this issue by implementing reactive path planning using potential fields.
A global map is built by the FastSLAM particle filter algorithm, which is introduced in \cite{montemerlo2007fastslam}.
The FastSLAM algorithm was run offline on a desktop computer with the data obtained during the flight.
We propose to connect localization, navigation, control, and mapping into one pipeline running online onboard of the UAV.
Our approach has the benefit of serving as a low-level no-maintenance layer for higher level algorithms for which it supplies the globally consistent map and the absolute position.

In the manuscript \cite{cui2014autonomous}, a SLAM solution aimed at autonomous navigation in forests is presented.
The laser scans from LIDAR are processed, clustering is applied, and clusters that are validated by a series of geometric descriptors are classified as features corresponding to trees.
These features are matched between successive scans to obtain a translation measurement, which is then fused with IMU estimate in a Kalman filter.
The fused position estimate tends to drift so a back-end GraphSLAM \cite{thrun2006graph} algorithm is implemented to detect loop closures with sufficient overlap between the current feature set and the previously visited feature sets.
The proposed method is backed by convincing experiments with both simulated dataset, and a real UAV platform.
Compared to raw scan matching techniques such as \cite{lu97imrp}, the feature-based approach is more robust to noise and outliers. 
Nevertheless, the geometric-based feature detector restricts the use for environments with tree-like features only.
Hence the technique cannot be used for a UAV deployed in another environment.

A combination of 2D LIDAR and IMU mounted on a spring is proposed in \cite{bosse2012zebedee}, where the oscillations of the springs modify the scan plane of the sensor, so that 3D data can be obtained.
It is a low-cost alternative to multi-channel 3D LIDARs that scan the environment with multiple spinning beams.
However, the weight of \SI{0.5}{\kilo\gram} is too heavy compared for the small UAVs considered within this paper.
Also, the oscillating mass could potentially destabilize the UAV.

A thorough experimental evaluation of scan registration algorithms for mapping purposes is presented in \cite{razlaw2015evaluation}. 
The authors designed a custom 3D rotating LIDAR to be able to obtain 3D laser scans of the environment.
The original ICP algorithm \cite{besl92icp} registers 3D geometrical shapes or points, which is improved in Generalized ICP \cite{segal2009gicp} to allow plane-to-plane registration.
In NDT \cite{magnusson2007ndt}, \cite{stoyanov2012ndt} the points are not registered directly.
First, a normal distribution grid map is modeled from one scan, then the second scan is registered into this model.
The performance of theses algorithms is measured in terms of Absolute Trajectory Error (ATE) \cite{sturm12iros} and Mean Map Entropy (MME) \cite{droeschel2014mme}.
While \cite{razlaw2015evaluation} evaluates the precision of alignment of laser scans or point clouds, which is an important metric, it does not guarantee that the method will be suitable for stabilizing a UAV carrying a lightweight 2D LIDAR.

Our proposed method is more focused on the ability to provide a solution even under non-ideal conditions such as low number of scan points and high outlier/inlier ratio.
Furthermore, the method needs to output the estimated motion in real time to achieve stability of the control feedback loop.
By addressing these common shortcomings we achieve a general, reliable method suitable for real-world applications, which is verified by numerous experiments in \refsection{sec:experiments}.




\section{2D lidar-based localization}
\label{sec:system}

\begin{figure}
  \centering
  \includegraphics[width=1.0\linewidth]{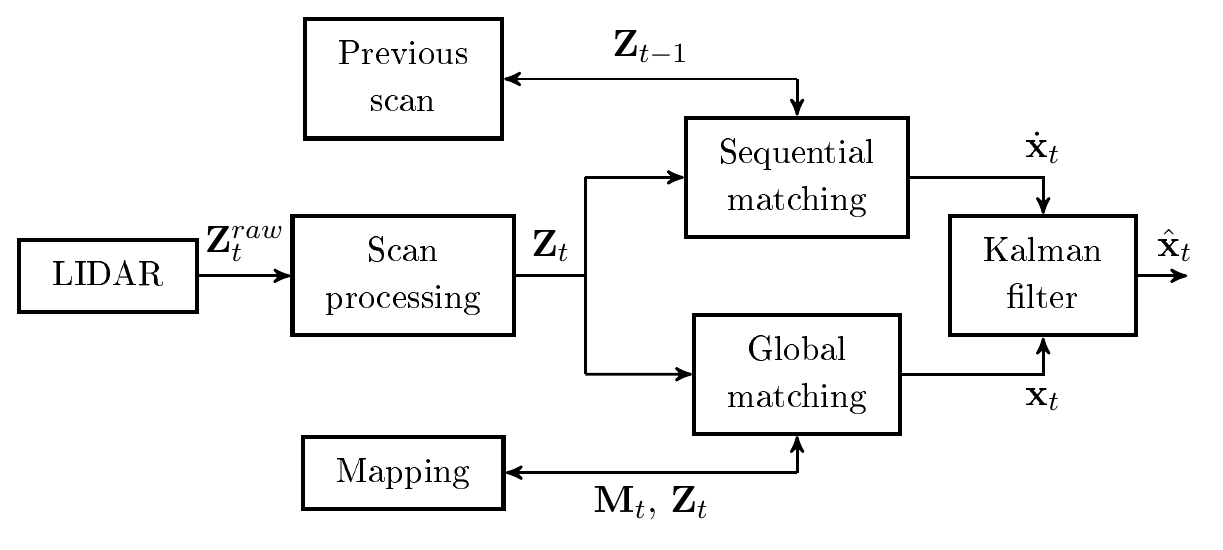}
  \caption[Localization pipeline diagram]{
    Diagram showing the sequence of modules in the localization pipeline.
  }
  \label{fig:localization_pipeline}
  \figvspace
\end{figure}

The proposed approach is divided into modules forming a pipeline illustrated in \reffig{fig:localization_pipeline}.

Each laser scan $\lscan^{raw}_t$ produced by the LIDAR mounted on the moving UAV is degraded by noise, outliers, and false detections.
These issues are addressed by the \module{Scan processing} module before scan matching so that the scans faithfully represent the surroundings.

Traditionally, the scan $\lscan_t$ is aligned to the previous scan $\lscan_{t-1}$ to obtain the displacement $\Delta\vect{x}$ between the two scans.
The UAV position cannot be determined reliably by directly integrating the relative displacements of successive scans as it is done by most of the systems proposed in literature, due to a drift accumulated from small errors in each scan matching.
Instead of directly integrating the relative displacements, we use the known timestamp of each scan to calculate the UAV velocity $\vect{\dot{x}}_t=\Delta\vect{x}/\Delta{t}$, where $\Delta{t}$ is the time difference between the two scans that are aligned in the \module{Sequential matching} module.
Simultaneously, $\lscan_t$ is also aligned by the \module{Global matching} module to the global map, which is gradually built in the \module{Mapping} module, to obtain the global position estimate $\vect{{x}}_t$.
Both estimates $\vect{x}_t$ and $\dot{\vect{x}}_t$ are fused in the \module{Kalman filter} to produce the final drift-free position estimate $\hat{\vect{x}}_t$.

\subsection{Scan processing}
\begin{figure}
  \centering
  \includegraphics[width=1.0\linewidth]{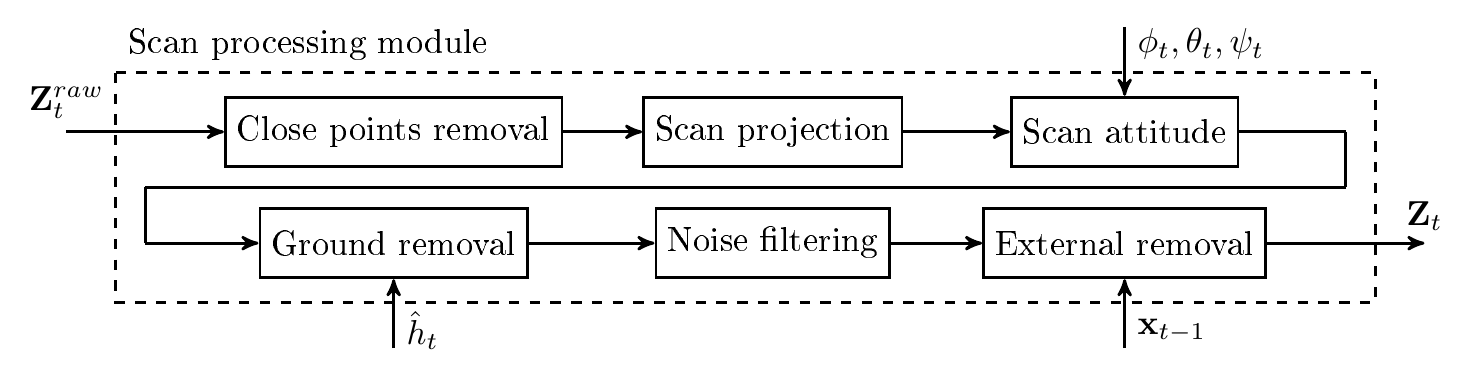}
  \caption[Scan processing pipeline diagram]{
    Submodules of the scan processing pipeline, with the raw laser scan $\lscan_t^{raw}$ on the input and processed scan $\lscan_t$ on the output.
    The scan attitude module has to be supplied by UAV roll, pitch, roll angles ($\phi,\theta,\psi)$, the ground removal module needs the altitude estimate $\hat{h}_t$, and the external removal module requires the last known position $\vect{x}_{t-1}$.
  }
  \label{fig:processing_pipeline}
  \figvspace
\end{figure}

Each scan $\lscan_t$ is preprocessed before $\vect{x}_t$ and $\dot{\vect{x}}_t$ can be estimated.
The sequence of operations performed on the laser scan is shown in \reffig{fig:processing_pipeline}.
The processed scan is a subset $\lscan_t = \lscan^{raw}_t \setminus ( \lscan^{close}_{t} \cup \lscan^{ground}_{t} \cup \lscan^{noise}_{t} \cup \lscan^{external}_{t} )$ converted from polar to Cartesian coordinates, and transformed by the UAV attitude and altitude.

The subset of close points $\lscan^{close}_{t} \subseteq \lscan^{raw}_t$ contains only points that satisfy the inequality $r < \uavsub{R}$, where $r$ is the range of the point in polar coordinates and $\uavsub{R}$ is the radius of the UAV.
In the case of the platform used for verification experiments shown in \reffig{fig:intro}, the radius of removed close points is $\uavsub{R}=\SI{0.395}{\metre}$ since the radius of the body frame is $\SI{0.275}{\metre}$ plus the radius of the propeller, which is $\SI{0.12}{\metre}$.
The points in $\lscan^{ground}_{t} \subseteq \lscan^{raw}_t$ correspond to laser beams that hit the ground.
After transforming the points by the UAV altitude and attitude, and converting to Cartesian coordinates, the ground points satisfy $p_z < h_{min}$, where $h_{min}$ is the ground threshold.
Setting $h_{min}$ to a higher value allows filtering out sloped terrain.
On the other hand, with higher $h_{min}$, more points will be removed from the already quite scarce scan.
Another subset $\lscan^{noise}_{t} \subseteq \lscan^{raw}_t$ contains the noise points that appear in the scans at random places due to the sensor imperfection.
Each point that has less than two neighbors in its radius belongs to this subset.
The last subset $\lscan^{external}_{t} \subseteq \lscan^{raw}_t$ contains points outside of the flight area.
These points could be highly dynamic and potentially ruin the estimation when there are more dynamic than static points present in the scans, hence it is safer to filter them out.

\subsection{Sequential matching}
The purpose of this module is to align two successive scans $\lscan_{t}$ and $\lscan_{t-1}$ in a way that minimizes the cost function that will be introduced later in \refsection{sec:scan_matching}.
The transformation, which has to be applied to acquire such alignment is also the transformation between the UAV poses $\vect{x}_{t}$ and $\vect{x}_{t-1}$ from which those two scans were obtained.
How the two scans are aligned is in detail described in \refsection{sec:scan_matching}.
The output of the matching method is a rigid transformation in 2D which is a combination of rotation $\varphi$ and translation $\left[T_x,\,T_y\right]$.

After the scans are aligned, and the final transformation between them is known, the current linear velocity of the UAV is obtained in the following way:
\begin{equation}
  \vect{\dot{x}}_t = \begin{bmatrix} T_x \\ T_y\end{bmatrix}\big/\Delta t,
\end{equation}
where $\Delta t$ is the time difference between scans $\lscan_{t-1}$ and $\lscan_{t}$ obtained from the timestamps which are filled every time a new scan is acquired from the sensor.
Similarly, the angular velocity is obtained as
\begin{equation}
  \dot{\varphi}_t = \varphi / \Delta t.
\end{equation}

\subsection{Global matching}
This module estimates the absolute position of the UAV in the flight area.
To achieve this goal, the current preprocessed scan $\lscan_t$ and the latest available map state $\map_{t-1}$ is required to localize the UAV.
The map can either be prepared in advance of built simultaneously during the flight.
The map $\map_{t-1}$ is cropped around $\vect{x}_{t-1}$ to the radius $r_c=1.2r_{max}$, with $r_{max}$ being the maximum range of the sensor.
As can be seen in \reffig{fig:church_map}, the built map is 3D, with cross-section varying at different heights.
To ensure correct alignment, only a vertical slice around the current flight height of the UAV is considered.

The cropping reduces the complexity of correspondence search and also limits the search to a small neighborhood which has the advantage that when there are two places in the map in which the crops are identical (such as corners formed by two walls), there is no ambiguity and the wrong match is not considered.

The actual scan alignment process is the same combination of state-of-the-art methods as in the sequential matching.
A detailed explanation resides in \refsection{sec:scan_matching}.
When scan matching of $\lscan_{t}$ to  $\map_{t-1}$ is finished, the module outputs the transformation matrix $\tf_t$.
The scan $\lscan_t$ is then transformed by the transform $\tf_t$ and current absolute position estimate is extracted from $\tf_t$.
If the UAV traveled more then \SI{0.5}{\metre} since the last map update, $\lscan_t$ is passed to the mapping module to get integrated into the global map.
The resulting estimated absolute position is passed to the Kalman filter to be fused with other measurements.
The global matching is run at \SI{1}{\hertz}, which is less than the frequency of sequential matching.
The reason is that the global matching estimate should correct the drift accumulated from the sequential matching.
The drift after \SI{1}{\second} is negligible, therefore, running the global matching more frequently does not bring any advantage and unnecessarily wastes CPU resources.

\subsection{Mapping}

\begin{figure*}
  \centering
      \begin{subfigure}{0.24\linewidth}
        \centering
\setlength{\fboxsep}{0pt}%
\setlength{\fboxrule}{1pt}%
\adjincludegraphics[width=1.0\textwidth,trim={{0.1\width} {0.1\height} {0.1\width} {0.1\height}}, clip]{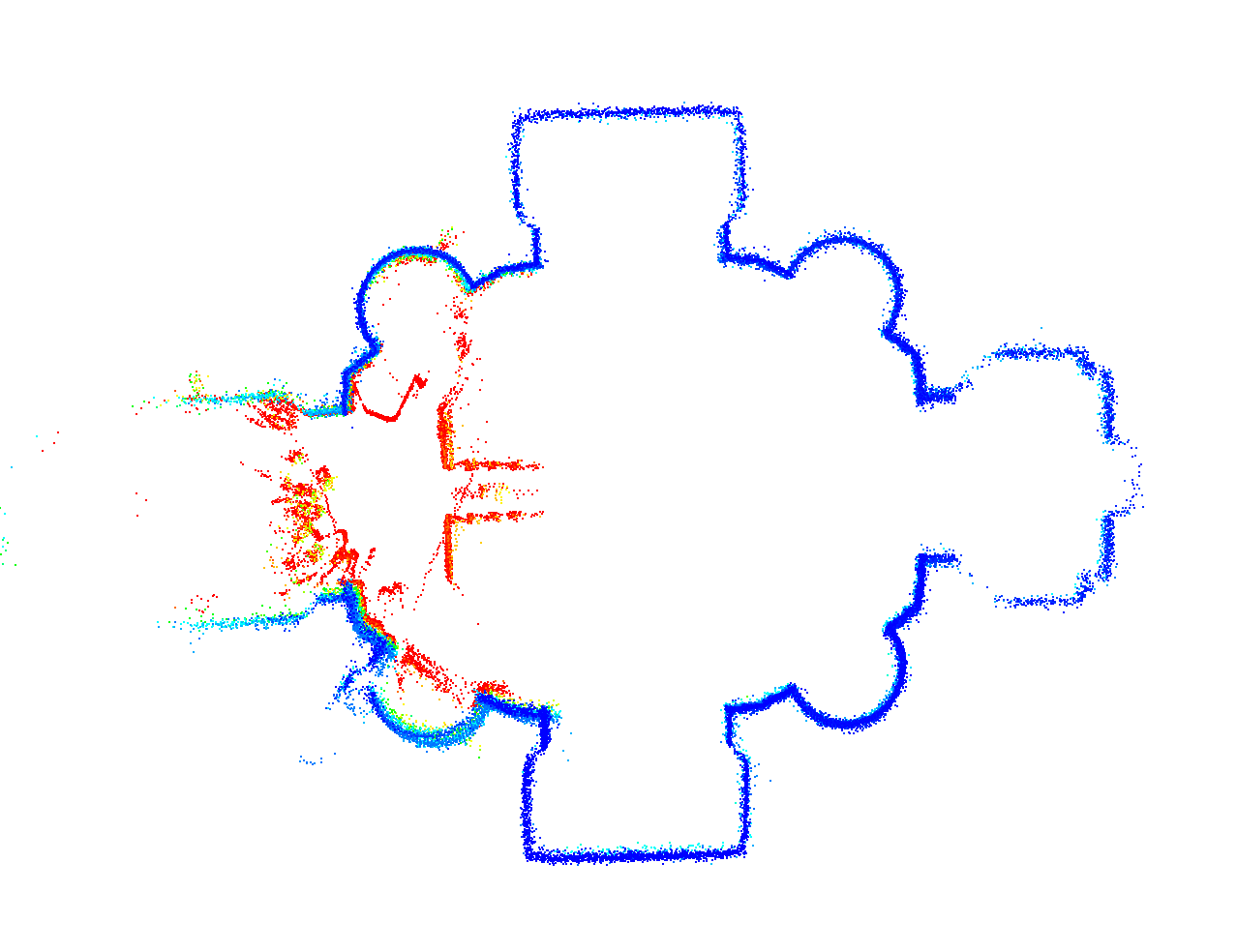}
      \end{subfigure}
      \begin{subfigure}{0.24\linewidth}
        \centering
\setlength{\fboxsep}{0pt}%
\setlength{\fboxrule}{1pt}%
\adjincludegraphics[width=1.0\linewidth, trim={{0.05\width} {0.05\height} {0.05\width} {0.05\height}}, clip=true]{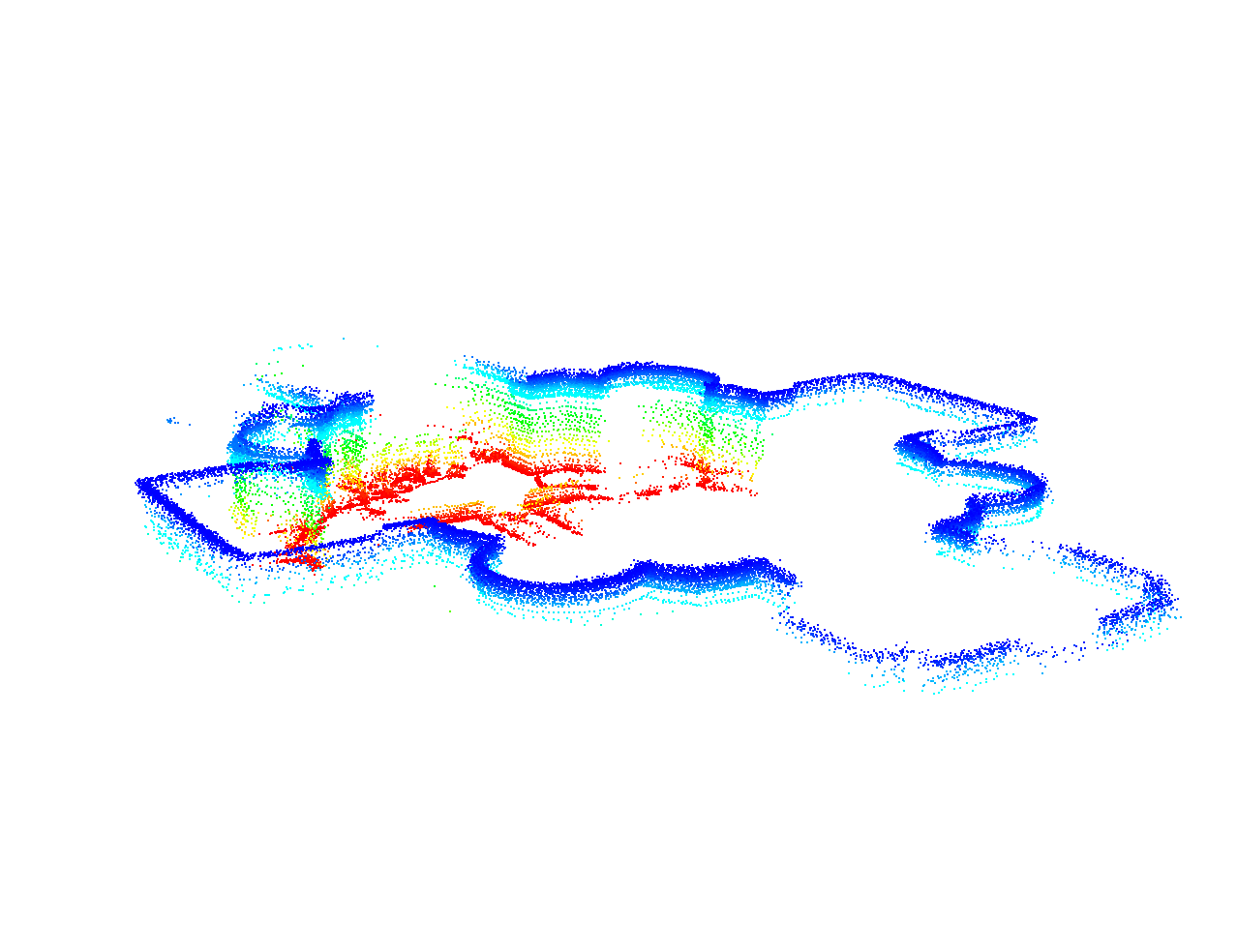}
      \end{subfigure}
      \begin{subfigure}{0.24\linewidth}
        \centering
\setlength{\fboxsep}{0pt}%
\setlength{\fboxrule}{1pt}%
        \adjincludegraphics[width=1.0\linewidth, trim={{0.1\width} {0.0\height} {0.1\width} {0.2\height}}, clip=true]{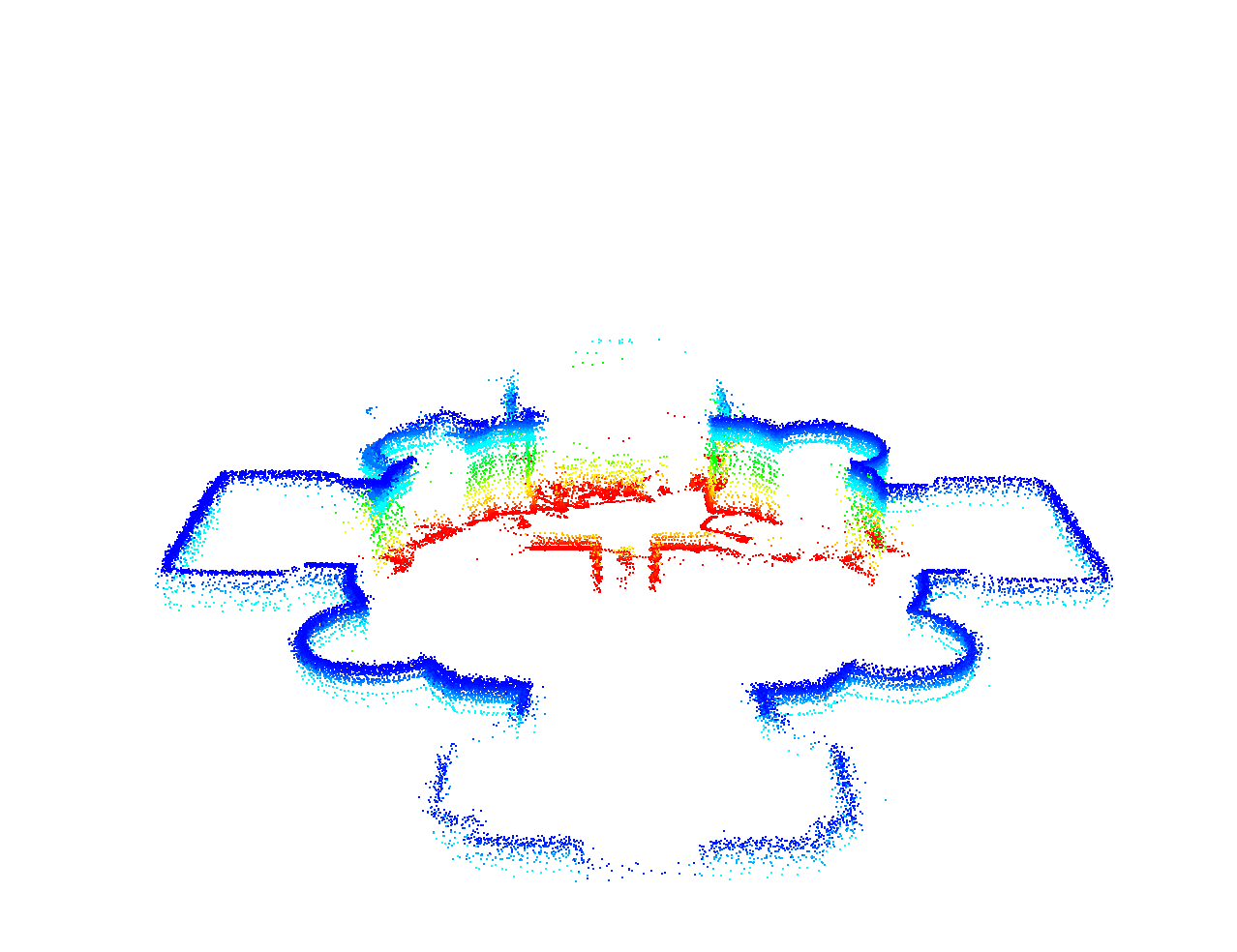}
      \end{subfigure}
      \begin{subfigure}{0.24\linewidth}
        \centering
\setlength{\fboxsep}{0pt}%
\setlength{\fboxrule}{1pt}%
        \adjincludegraphics[width=1.0\linewidth, trim={{0.05\width} {0.0\height} {0.1\width} {0.15\height}}, clip=true]{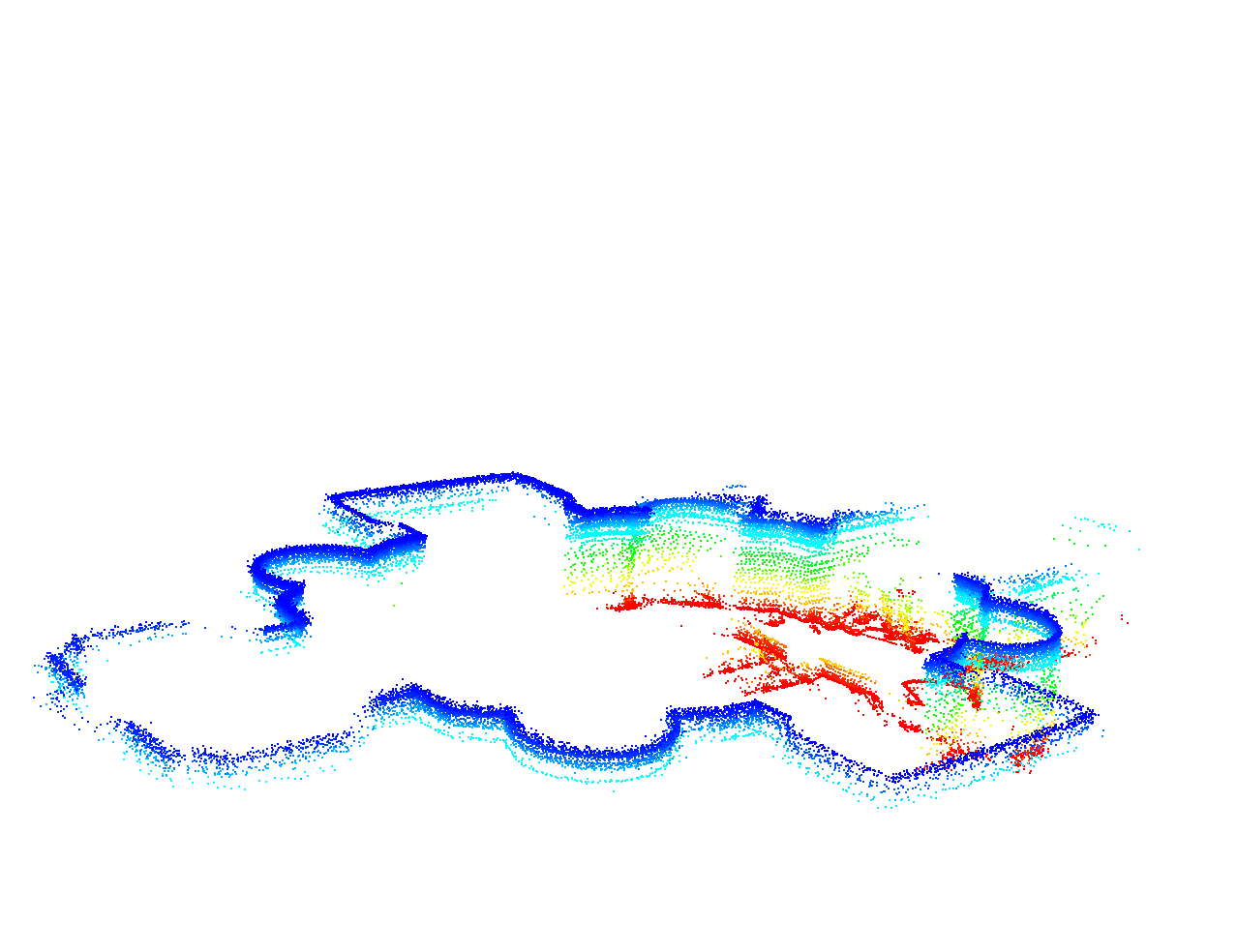}
      \end{subfigure}
  \caption{The 3D model of the Stara Voda church as scanned by the LIDAR and aligned online onboard the UAV by our scan matching algorithm.
    The colors represent the height of captured points - red for lowest points, blue for highest points.
    \label{fig:church_map}
  }
  \figvspace
\end{figure*}

The map is stored in the octree data structure \cite{hornung13auro} with resolution constrained to $r=\SI{0.2}{\metre}$.
A new scan is integrated into the map every second, adding only points that satisfy $\lVert \vect{p_i} - \mu(o_j) \rVert^2 \ge r$ for all octree cells $o_j\in\mathcal{O}$ with $\mu$ being the center of the cell.
When queried for the current state of the map, the module generates a point cloud representation $\map$ with points $\vect{p}_j = \mu(o_j)$ for $j = 1,\ldots, |\mathcal{O}|$. 
The mapping module also keeps a dense point cloud composed from all successfully aligned scans that can be used as a side output of UAV missions in an unknown environment to provide an overview for end-users.
Such obtained 3D model of the church in Stara Voda can be seen in \reffig{fig:church_map}.

\subsection{Kalman filter}

The outputs of sequential and global matching are fused as measurements of velocity and position respectively in a linear Kalman filter.
Since the components of the UAV position $\vct{r}=[x,y]^T$ are independent on each other, the estimation can be decoupled for each axis, which decreases the size of the state-space description.
The filters for $x$ and $y$ axes have an identical model, and only $x$-axis estimation will be discussed for brevity.

The state vector for x-axis is defined as $\vct{x}=(x,\dot{x},\ddot{x})$, and the discrete formulation of the dynamic LTI stochastic system is
\begin{align}
  \vct{x}_{t+1} &= \m{A} \vct{x}_t + \m{B}\vct{u}_t + \vct{v}_t, \\
  \vct{y}_{t} &= \m{C} \vct{x}_t + \m{D}\vct{u}_t + \vct{e}_t,
\end{align}
where $\vct{x}_t$ is the state vector, $\vct{u}_t$ is the input vector in the sample $t$.
The process noise $\vct{v}_t$ and measurement noise $\vct{e}_t$ describe the stochastic part of the system.
The noise is drawn from the normal distributions $\vct{v}_t\sim\mathcal{N}(\vct{0},\m{Q})$ and $\vct{e}_t\sim\mathcal{N}(\vct{0},\m{R})$ respectively, i.e., zero-mean Gaussian noise is assumed.
The rotation around the world-fixed z-axis is estimated using a two-state model $\boldsymbol{\varphi}=(\varphi,\dot{\varphi})$.

Since the proposed method is developed for UAVs with 2D LIDARs, the altitude or height of the UAV above ground is unobservable from the LIDAR scans only.
To estimate the UAV height, another filter with the three-state model $\vct{z}=(z,\dot{z},\ddot{z})$ is used.
The gravity-aligned acceleration measured by the IMU drives the prediction step of the filter.
When navigating enclosed spaces, the height above terrain is more informative than e.g., above-sea-level altitude. 
Consequently, the measurements from a downward facing laser rangefinder are fused in the filter as corrections. 
Additional correction consisting of the rate of change of barometric altitude is applied to the velocity state to mitigate high-innovation changes in state estimate when briefly flying above an obstacle.




\section{Scan matching}
\label{sec:scan_matching}

The core of the proposed solution is a method that estimates the difference $\deltapos$ between UAV positions $\posref$ and $\posact$ at time $t_1,\,t_2\inr^+$ reliably in arbitrary environment with obstacles.
The problem of finding $\deltapos$ corresponds to finding the rigid 2D transformation
between laser scans $\lscan_{t_1}$ and $\lscan_{t_2}$ that were taken at positions $\posref$ and $\posact$ respectively.
This section presents the details of the implemented scan matching algorithm that finds $\tf$ (initialized to the identity matrix) by aligning $\lscan_{t_2}$ to $\lscan_{t_1}$ as follows:

\begin{enumerate}
  \item Establish correspondences between points from $\lscan_{t_2}$ and $\lscan_{t_1}$ such that
    \begin{equation}
      c(\pr_i)=\operatorname*{argmin}_{\ps_j \in \lscan_{t_2}} \lVert \pr_i - \ps_j \rVert^2,
    \end{equation}
    where $\pr_i$ is the $i$-th point from the laser scan $\lscan_{t_1}$
    and $\ps_j$ the $j$-th point from $\lscan_{t_2}$ with
    $i\in\left\{1,\ldots, |\lscan_{t_1}|\right\}$
    and $j\in\left\{1,\ldots,|\lscan_{t_2}|\right\}$.
    The operator $|\lscan|$ is the number of points in scan $\lscan$.
  \item Compute the transformation that aligns the laser scans by minimizing the error function.
    \begin{equation}
      E(\tf)=\sum_{i=1}^{n}\lVert \pr_i - \tf c(\pr_i) \rVert^2.
    \end{equation}
  \item Apply the obtained transformation to every $\ps_j \in \lscan_{t_2}$.
  \item Evaluate the quality of alignment and iterate all steps until a stopping condition is met.
\end{enumerate}

The original ICP algorithm is not suitable for aligning scans with small overlaps and a high number of outliers.
These properties are however typical for most of the scans coming from a moving UAV equipped with a slowly rotating LIDAR that outputs a relatively small amount of points.
The Scanse Sweep LIDAR used during testing of the stabilization system generates scans at \SI{5}{\hertz}.
Each scan contains less than 200 points.
The algorithm was thus modified to allow precise and reliable alignment of non-ideal scans.
Analysis of each step of the proposed approach follows.

\begin{figure}
  \centering
  \includegraphics[width=1\linewidth]{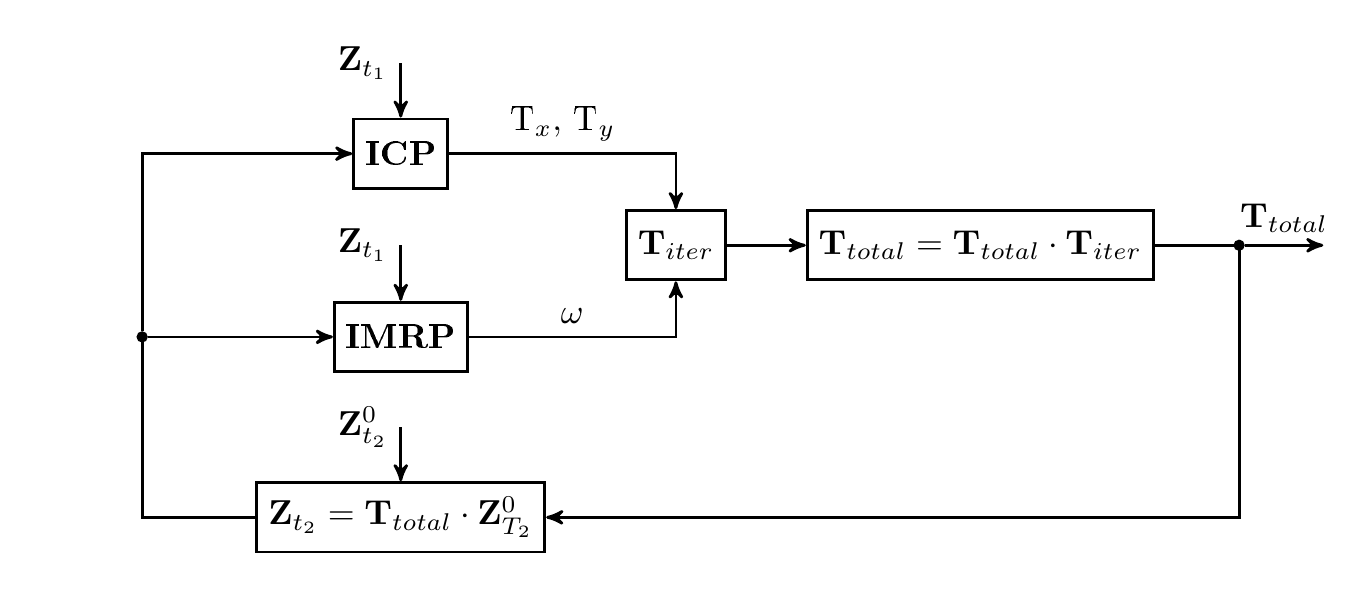}
  \caption[Scan matching pipeline diagram]{
    A transformation $\tf_{iter}$ is found by combining translation $\text{T}_x,\,\text{T}_y$, from the ICP algorithm, with the rotation $\omega$ obtained from IMRP method.
    Then the total transformation $\tf_{total}$ is updated and applied to $\lscan_{t_2}^0$ to obtain $\lscan_{t_2}$ which is used to estimate $\tf_{iter}$ in the next iteration.
  }
  \label{fig:matching_pipeline}
  \figvspace
\end{figure}

\subsection{Establishing correspondences}

Finding the appropriate correspondences is critical to correct scan alignment.
The rate of convergence (number of iterations until reaching minimum), robustness to noise and outliers, and the time complexity all depend on the choice of correspondences.

The information carried by high-density scans is often redundant, and thus can benefit from finding distinctive features that represent well the geometry of the environment with much less data points.
The state-of-the-art algorithms \cite{zhang2014loam}, \cite{shan2020lio} that are tailored for multi-channel LIDARs use these geometric features in correspondence search and minimization of their distance.
While feature extraction can provide a unique representation of the environment, with low-density scans it cannot find sufficient amount of features to estimate the transformation between scans reliably.
Point-to-point correspondences are more appropriate for a sensor with low sampling rate since the measured point cloud is rather sparse and obtaining unambiguous geometric features is not possible due to insufficient detail of scanned objects.
Since the point-to-point correspondence search does not discard any points, no spatial information is lost in the process, and the transformation can still be estimated even in situations where feature-based methods fail.

\subsubsection{Linear interpolation}
In the original ICP algorithm \cite{besl92icp}, as it is used in state-of-the-art literature, the closest point correspondences are used.
The correspondence $c(\ps_j)$ of point $\ps_j \in \lscan_{t_2}$ is found as the closest point $\pr_i \in \lscan_{t_1}$ in terms of Euclidean distance.

\begin{figure}
  \centering
  \includegraphics[width=0.7\linewidth]{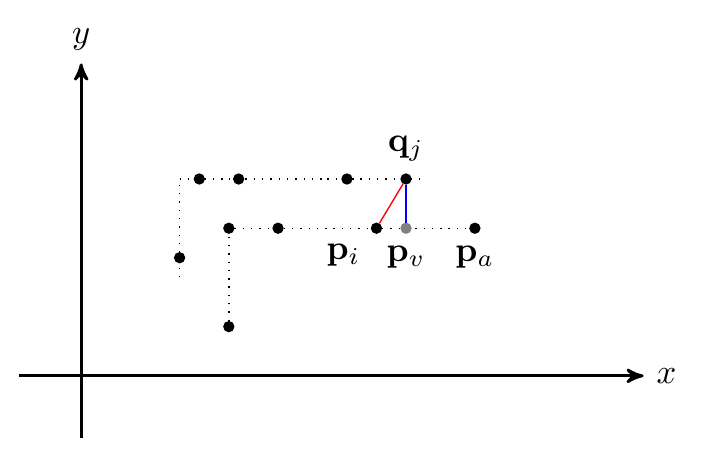}
  \caption[Correspondence search at line segments]{
    The correspondence search on the line segments.
    Red line represents the correspondence found by regular ICP, i.e., the closest point $c(\ps_j)=\pr_i$.
    Searching for the correspondence on a line segment produces a virtual point $\pr_v$ lying between the closest point $\pr_i$ and the adjacent point $\pr_a$.
    The correspondence then becomes $c(\ps_j)=\pr_v$.
  }
  \label{fig:line_segment}
  \figvspace
\end{figure}

To achieve the accuracy of matching, in sparse point clouds, we propose using the linear interpolation of $\lscan_{t_1}$.
The neighbor point of $\pr_i$ that is closer to $\ps_i$ is selected to be the adjacent point $\pr_a$.
A virtual correspondence $\pr_v$ is formed on the closest point of the line segment $\overline{\pr_i\pr_a}$ as shown in \reffig{fig:line_segment}.

\subsubsection{Iterative matching range point}
As mentioned in \cite{lu97imrp}, the Euclidean distance metric leads to fast convergence of the components $T_x,\,T_y$ of the transformation between two scans.
However, the convergence of the rotational component $\omega$ is much slower, which is to be expected, as the Euclidean distance metric of two points yields no explicit information about the rotation.
As a result, the algorithm tries to minimize the rotational error by translation rather than rotation, thus causing an unnecessary decrease in the rate of convergence of the whole algorithm, which is a problem when the matching is constrained by the real-time requirement.

Similarly as \cite{lu97imrp}, we use the Iterative matching range point (IMRP) to improve the performance of rotation estimation.
The structure of the algorithm is the same as of ICP, and the only difference is that the correspondence search uses a rotation-based metric.
The correspondences are searched in the polar coordinates in which the laser scans are natively represented.
For every point $\ps_j=(r_j,\,\varphi_j)\in \lscan_{t_1}$, we find the corresponding point $\pr_i=(r_i,\, \varphi_i) \in \lscan_{t_2}$ that satisfies $\varphi_i \in \left [ \varphi_j - B_{\omega},\, \varphi_j + B_{\omega} \right ]$ and minimizes the metric $\lVert r_j-r_i \lVert^2$, where $B_{\omega}$ is the maximal expected rotation error.
The expected maximal rotation error $B_{\omega}$ is subject to change in every iteration as the scan alignment converges.
Since the rotational residual decreases exponentially with the number of iterations, we set $B_{\omega}(t) = B_{\omega}(0)e^{-\alpha t}$ to prevent incorrect correspondences in later iterations.
The constant $\alpha=0.03$ was used in all experiments.

\subsubsection{Iterative dual correspondences}
Scan matching based on IMRP converges faster in the rotational component. On the other hand, the convergence of translation is slower than in ICP.
To obtain the best possible estimate, we use ICP correspondences for estimating the translation and IMRP correspondences for rotation estimation as illustrated in \reffig{fig:matching_pipeline}.
This approach is referred to as Iterative dual correspondences (IDC) in \cite{lu97imrp}.

\subsubsection{Outlier rejection}
The found correspondences typically contain a large number of outliers.
The points measured from position $\vect{x}_{t_1}$ that cannot be seen from position $\vect{x}_{t_2}$ due to moving out of range of the sensor or occlusion do not have a correct correspondence pair and have to be rejected to prevent the incorrect alignment.

Robustness to outliers is improved by incorporating the fraction of inliers into the distance function \cite{jeff06frmsd} thus forming a new distance measure Fractional Root Mean Squared Distance (FRMSD) which is defined as
\begin{equation}
  \text{FRMSD}\left ( \lscan_{t_1},\,\lscan_{t_2},\,f,c \right ) = \frac{1}{f^{\lambda}}\sqrt{\frac{1}{f\cdot n} \sum_{\point\in \lscan_{t_2}}\lVert\point-c(\point)\rVert^2},
\end{equation}
where $f\in \left [0,\,1 \right ]$ is the fraction of $n$ points in $\lscan_{t_2}$ sorted by $\lVert\point-c(\point)\rVert^2$ that will be used for alignment.

The parameter $\lambda$ depends on the expected outlier density, and according to \cite{jeff06frmsd} should be set to $\lambda=1.2$ for general-purpose scan matching in two dimensions.
By fine-tuning $\lambda$, one can adjust the ratio of false positives to false negatives.
A smaller value of $\lambda$ can reject correct correspondences as outliers while on the other hand, higher values result in more outliers classified as inliers.

\subsubsection{Weighting}
Even with the outliers filtered out, correspondences with greater distance still contribute to the estimated transform more than correspondences separated by a smaller gap.
The performance of the matching is improved by assigning a weight to each correspondence pair as:
\begin{equation}
  w(d_i)=1-\frac{d_i}{\text{max}(d_{1,\,\ldots,\,n})},
\end{equation}
where $w(\cdot)$ is the weighting function and $d_i$ the distance of i-th correspondence pair.

\subsection{Computing transformation}
Once the correspondences have been established, we can compute the registration, i.e., the transformation that minimizes the cost function defined as the sum of squared residuals
\begin{multline}
  E(\tf)=\sum_{i=1}^{n} \bigg( \left( x_i \cos\omega - y_i \sin\omega + T_x -x'_i \right)^2 \\
  + \left( x_i \sin\omega + y_i \cos\omega + T_y -y'_i \right)^2 \biggr).
\end{multline}
Minimizing $E(\tf)$ we get
\begin{align*}
  \omega &= \arctan \frac{S_{xy'}-S_{yx'}}{S_{xx'}+S_{yy'}}, \\
  T_x &= \mu'_x - \left( \mu_x \cos \omega - \mu_y \sin \omega \right), \\
  T_y &= \mu'_y - \left( \mu_y \sin \omega + \mu_y \cos \omega \right),
  \numberthis
\end{align*}
with $\mu$ being the weighted mean and $S_{ab'}$ the weighted covariance defined as
\begin{align*}
  S_{ab'} &= \sum_{i=1}^{n}w_i (a_i-\mu_a)(b'_i-\mu'_b).
  \numberthis
\end{align*}

\subsection{Evaluation and iteration}

The solution convergence is checked by the difference of FRMSD between the current and previous iteration.
Since the error typically decreases exponentially, the difference between the consecutive errors is compared against a close-to-zero constant and if the difference is smaller, the algorithm has converged and further iterations are not necessary.
Similarly, when the FRMSD is less than $\SI{1}{\centi\metre}$ the matching does not continue.
The time available for matching is set to $\SI{50}{\milli\second}$ to satisfy the real-time requirement of localization.
When the time limit is exceeded the matching will be stopped, regardless of the current error.



\section{Experimental verification}
\label{sec:experiments}

\begin{figure}
  \centering
  \begin{tikzpicture}
    \node[anchor=south west,inner sep=0] (a) at (0,0) {\adjincludegraphics[width=0.5\linewidth, trim={{0.0\width} 0 {0.0\width} 0}, clip]{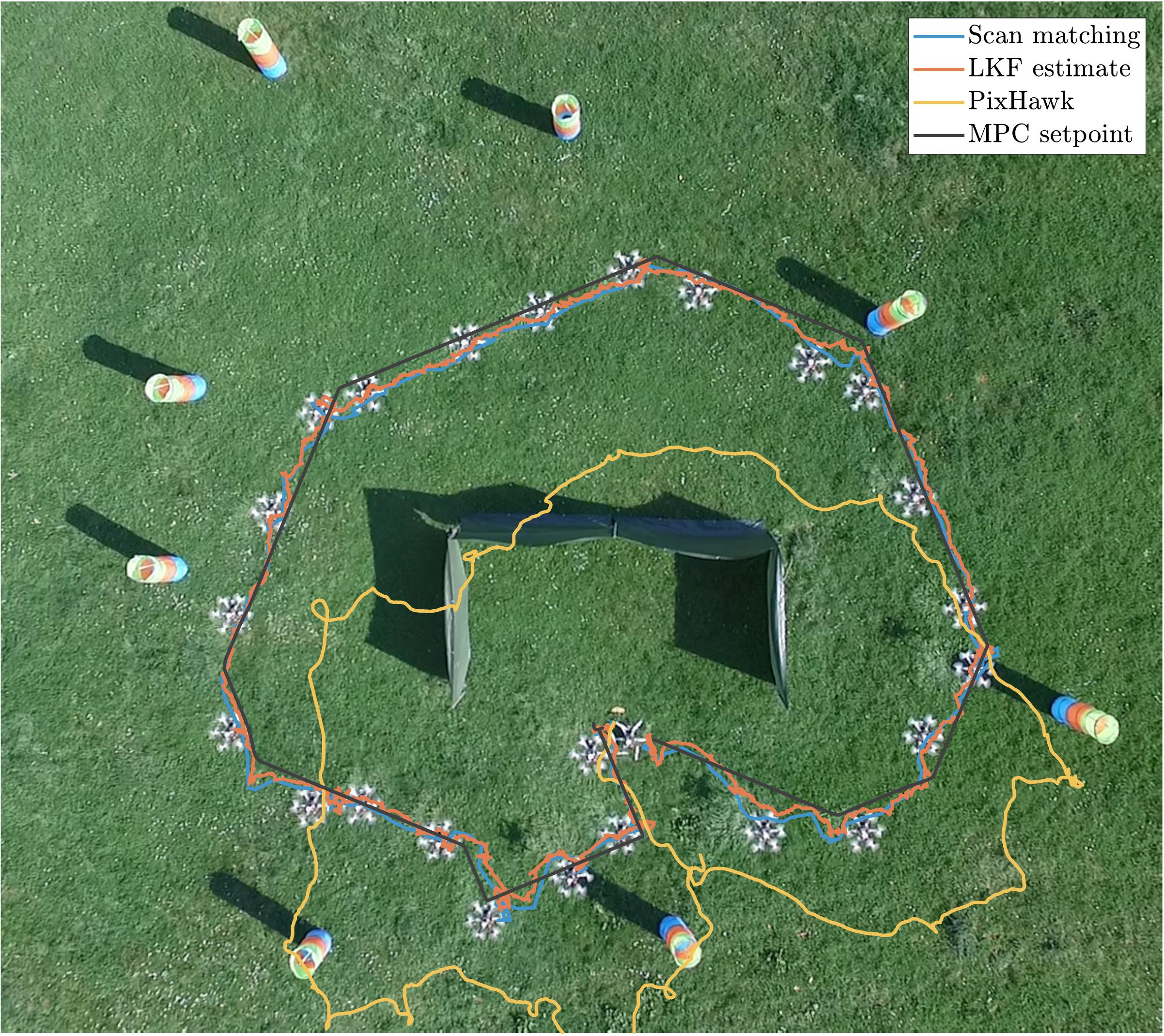}

      \adjincludegraphics[width=0.5\linewidth, trim={{0.08\width} 0 {0.0\width} 0}, clip]{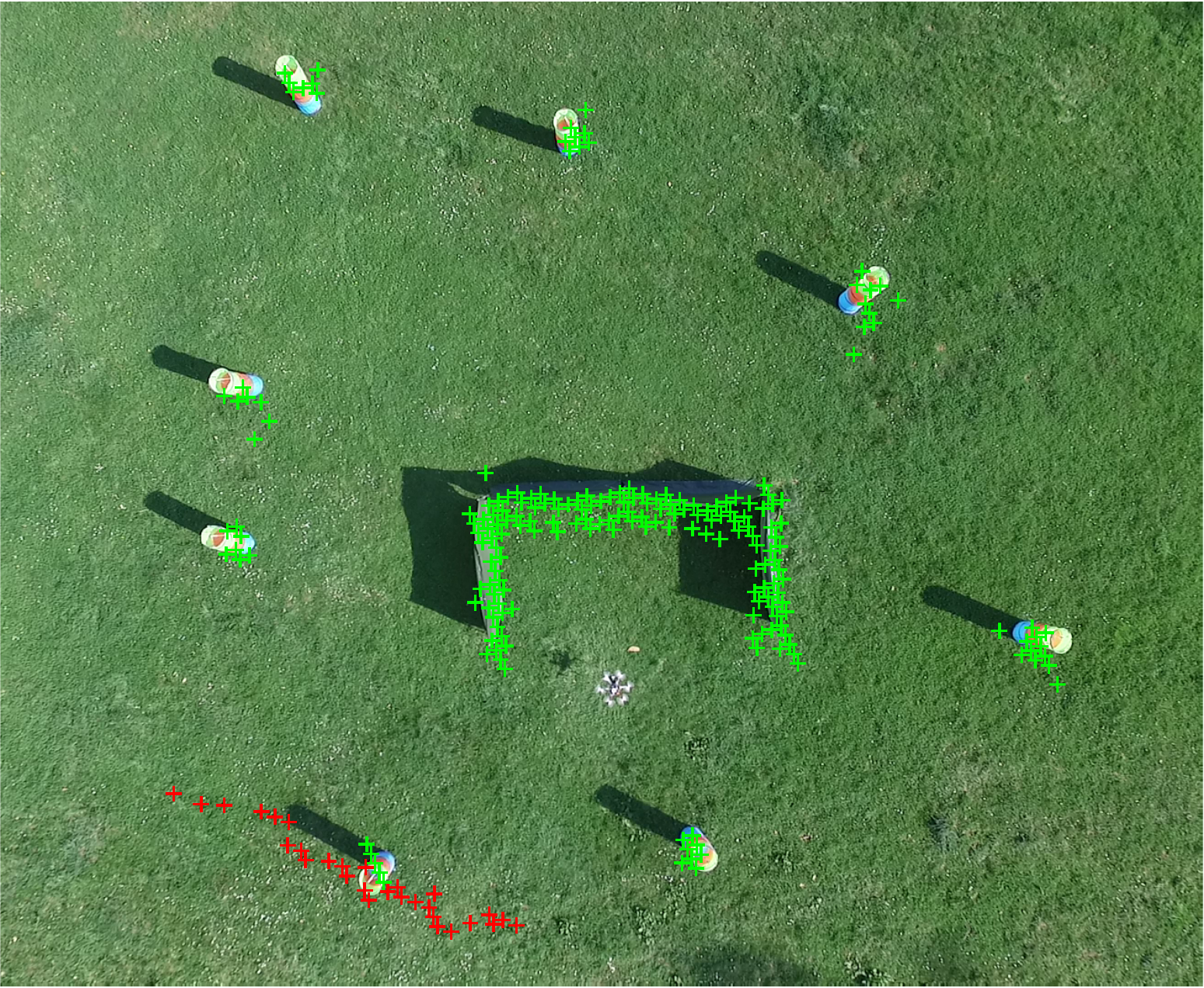}};
    \begin{scope}[x={(a.south east)},y={(a.north west)}]

      \newcommand\rectx{0.25}
      \newcommand\recty{0.77}
      \newcommand\rectxx{0.485}
      \newcommand\rectyy{0.98}
      \newcommand\lx{0.28}
      \newcommand\ly{0.95}
      \newcommand\lxd{0.05}
      \newcommand\lyd{0.05}
      \newcommand\textpos{2.5}

      \filldraw[fill=white, draw=black] (\rectx,\recty) rectangle (\rectxx,\rectyy);
      \draw[color={rgb:blue,3;green,2;white,2}] (\lx,\ly) -- (\lx+\lxd,\ly) node[color=black,align=left,anchor=west]{\tiny Scan matching};
      \draw[color={rgb:red,3;green,1;white,2}] (\lx,\ly-1*\lyd) -- (\lx+\lxd,\ly-1*\lyd) node[align=left,anchor=west,color=black]{\tiny Kalman filter};
      \draw[color={rgb:red,1;yellow,3;white,1}] (\lx,\ly-2*\lyd) -- (\lx+\lxd,\ly-2*\lyd) node[color=black,align=left,anchor=west]{\tiny GNSS};
      \draw[color={black}] (\lx,\ly-3*\lyd) -- (\lx+\lxd,\ly-3*\lyd) node[color=black,align=left,anchor=west]{\tiny Setpoint};

    \end{scope}

  \end{tikzpicture}
  \caption{The left image is stitched from a video with the UAV position snapshot taken every 10 seconds during the outdoor stabilization experiment.
  The crosses on the right image represent the final map with outliers depicted in red.}
  \label{fig:stitch_top_view_slam}
  \figvspace
\end{figure}
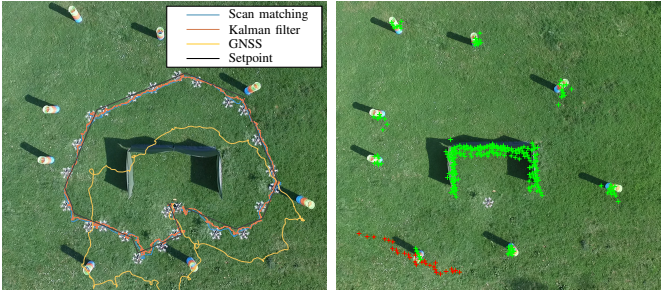


\begin{figure*}
  \begin{tikzpicture}
    \node[anchor=south west,inner sep=0] (a) at (0,0) {
        \adjincludegraphics[width=0.245\linewidth, keepaspectratio, trim={{0.05\width} 0 {0.1\width} 0}, clip]{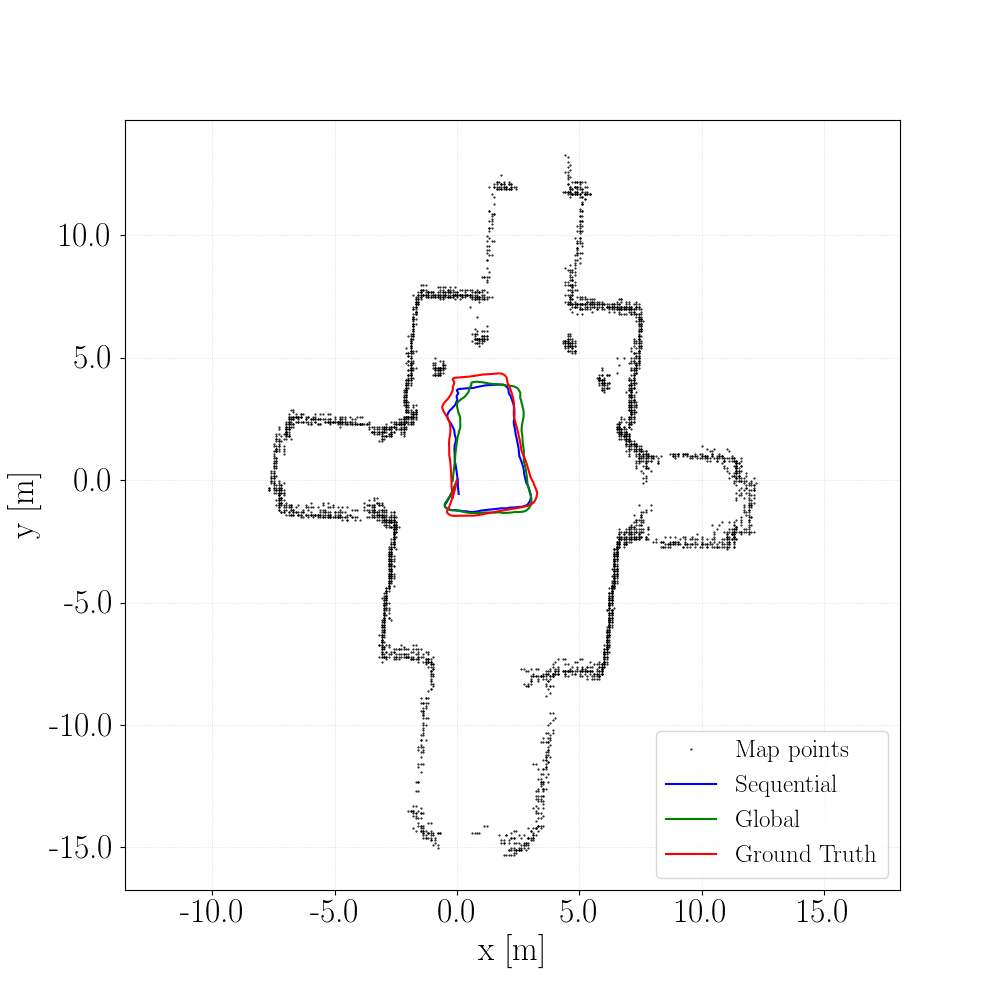}
        \adjincludegraphics[width=0.245\linewidth, keepaspectratio, trim={{0.05\width} 0 {0.1\width} 0}, clip]{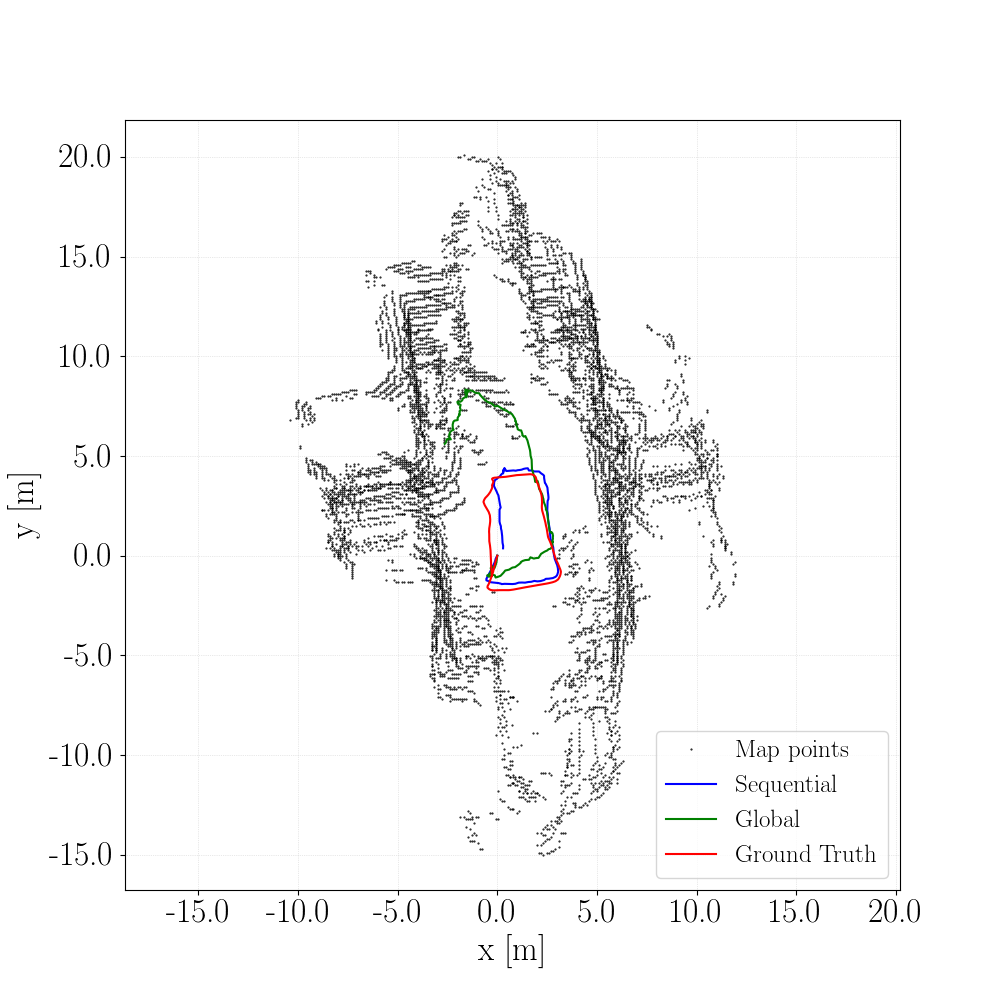}
        \adjincludegraphics[width=0.245\linewidth, keepaspectratio, trim={{0.05\width} 0 {0.1\width} 0}, clip]{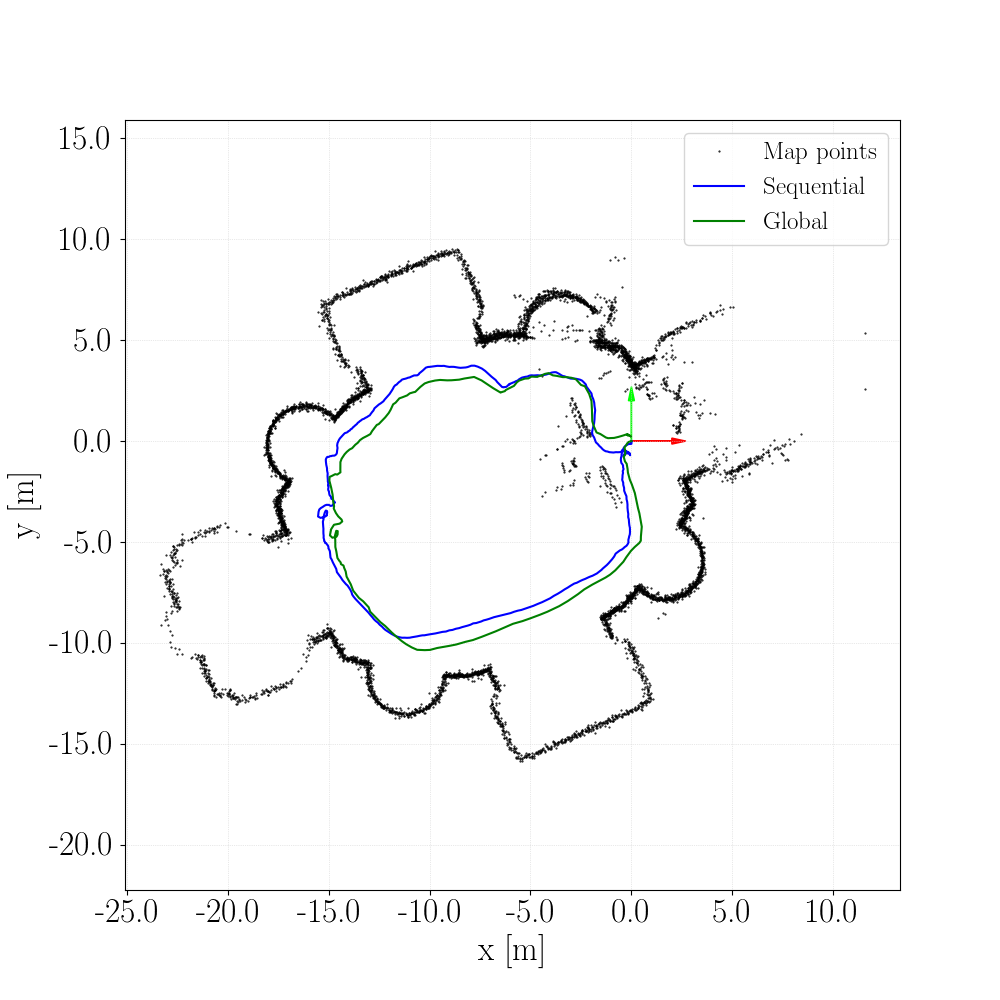}
        \adjincludegraphics[width=0.245\linewidth, keepaspectratio, trim={{0.05\width} 0 {0.1\width} 0}, clip]{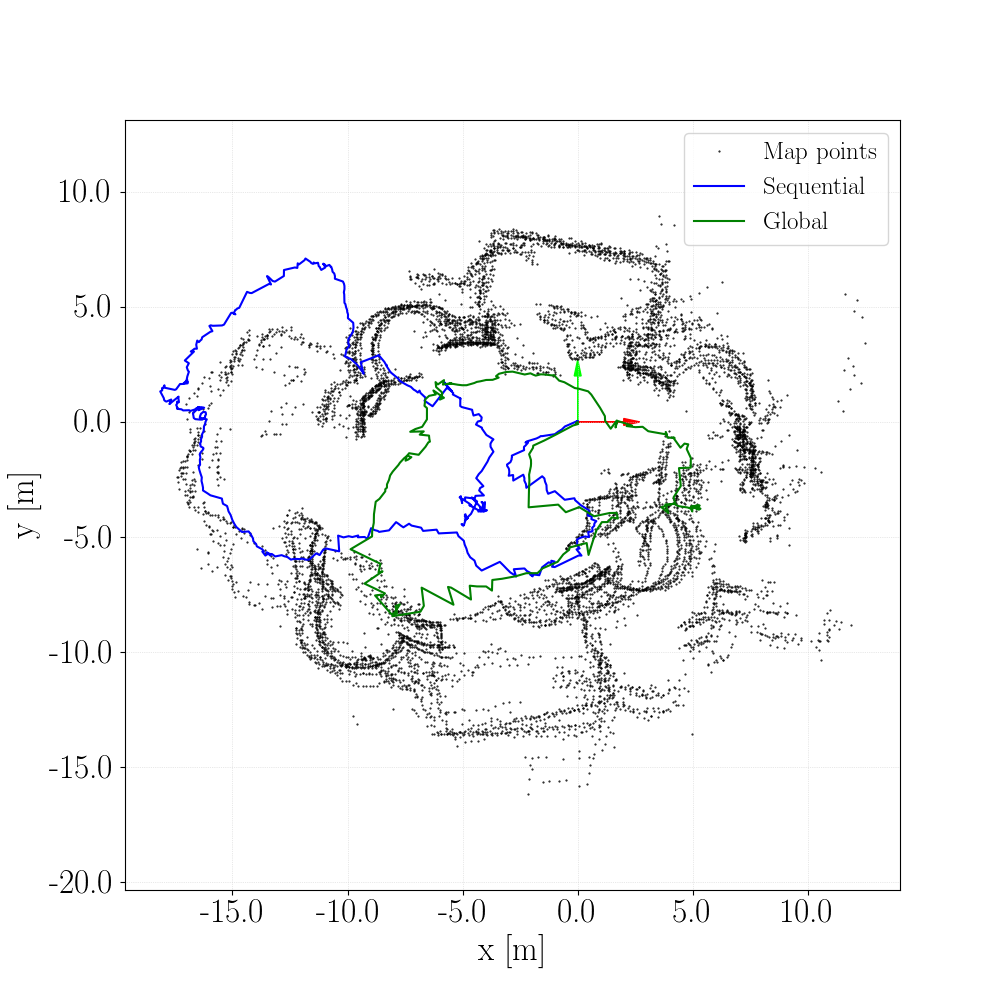}
    };

  \end{tikzpicture}
  \caption{The sequential (blue) and global (green) trajectories were obtained during a rectangular flight in the Chlumin church (first two plots) and the Stara Voda church (last two plots). 
    The ground truth trajectory (red) from tracking total station is included in the \textit{Chlumin dataset} to evaluate the performance of the method.
    A global map represented by a point cloud was gradually built online.
    The first and third plots show results of our proposed localization method, the second and fourth plots are obtained using the unmodified state-of-the-art ICP.
  \label{fig:chlumin_top}
}
  \figvspace
\end{figure*}



\subsection{Outdoor stabilization experiments}
\label{sec:experiments_hw}

The first experimental setup was realized at an outdoor experimental site with obstacles to prove that the localization system can stabilize the UAV even in a difficult environment with a low number of points in the laser scans (mean number of points in each scan was 49 after preprocessing).
Although our method was developed primarily for indoor usage, this outdoor experiment proves the robustness of LIDAR-based solution in direct sunlight, which is a situation, where camera-based methods can fail.
The Scanse Sweep LIDAR was used in this experiment.
The UAV was commanded to follow a trajectory that ends close to the takeoff location to give the localization system the possibility of loop closure.
The obtained map and estimated trajectory with GNSS trajectory for comparison are shown in \reffig{fig:stitch_top_view_slam}.
The trajectories do not match due to the GNSS estimate drifting a few meters after takeoff.
A small but noticeable control error can be observed as the setpoint was not tracked precisely, which could be fixed by increasing the controller position gain.

\subsection{Chlumin church evaluation}
A dataset containing ground truth pose trajectory was prepared to evaluate the precision of the localization algorithm.
The recording of the dataset took place in the Chlumin church.
Our experimental UAV platform was equipped with lightweight RPLIDAR A3 sensor, and was flown manually around the inside walls by an experienced pilot.
The ground truth was obtained from the Leica Nova MS60 total station \footnote{Leica Nova MS60, \url{https://w3.leica-geosystems.com/downloads123/zz/tps/nova_ms60/brochures-datasheet/Leica\%20Nova\%20MS60\%20DS_en.pdf}} that tracks the UAV using a laser beam reflected by a prism carried by the tracked UAV.
\reffig{fig:chlumin_top} shows both sequential and global trajectories compared to the ground truth poses.
The evaluation of the performance of the developed localization method was split into sequential performance and global performance, since each step of our system affects the local and global matching differently.

\begin{figure}
  \centering
  \includegraphics[width=1.0\linewidth]{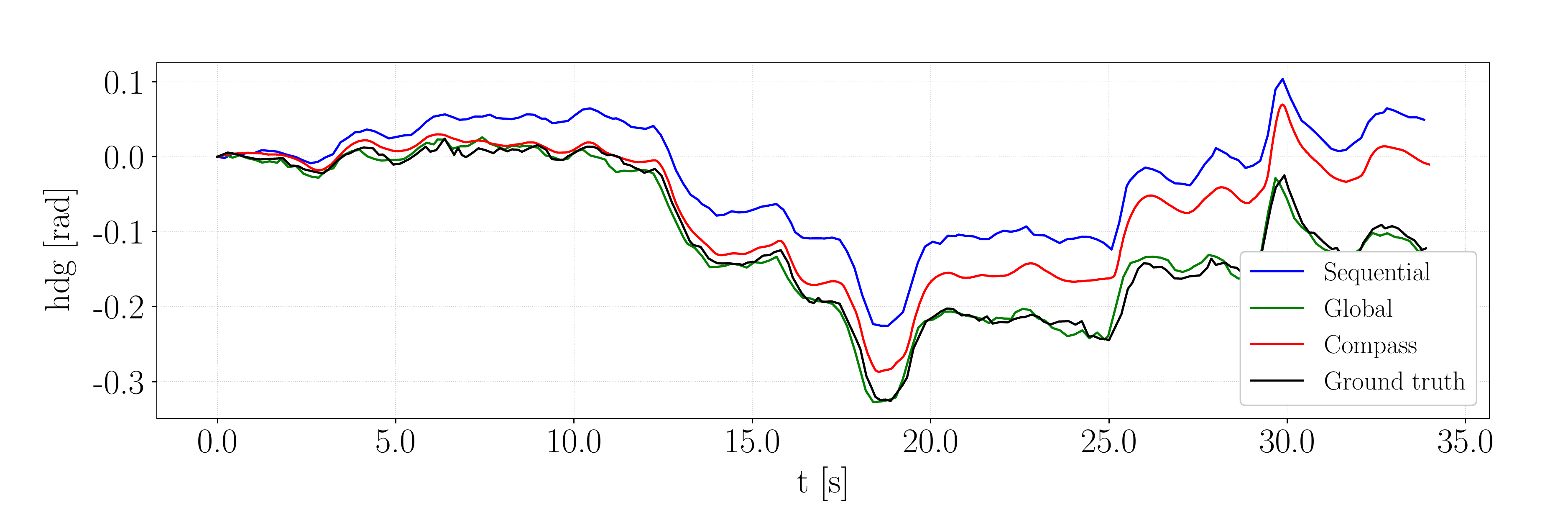}
  \caption{
    The sequential estimate of heading trajectory from the \textit{Chlumin dataset} drifts over time.
    By aligning scans into global map, we can obtain heading estimate that has lower error than compass.
  \label{fig:heading}
  }
  \figvspace
\end{figure}

We quantified the localization performance by the absolute trajectory error (ATE) \cite{sturm12iros} with respect to the ground truth trajectory obtained from the tracking total station.
\reffig{fig:ate} depicts the ATE of sequential and global trajectories.
Since the ATE represents mainly the translational components of the error (although heading errors can manifest themselves as a wrong translation) we include also the RMSE of the estimated heading.
The trajectories of the heading as estimated by sequential and global matching are compared to compass and ground truth in \reffig{fig:heading}.
Additionally we calculated the map mean entropy (MME) \cite{droeschel2014mme} as 
\begin{equation}
  H\left(\map\right)=\frac{1}{\left|\map\right|}\sum^{\left|\map\right|}_{k=1}\frac{1}{2}\ln |2\pi e \Sigma(q_k)|,
\end{equation}
where $\Sigma(q_k)$ is the sample covariance of mapped points in local radius $r$ around map point $q_k$.
The correlation between the ATE and the sharpness of the map represented by the MME can be used to assess position estimate quality of the \textit{Stara Voda dataset} even in the absence of ground truth.
The \reftab{tab:chlumin} shows the results of the evaluation of our modified ICP.


\begin{table}[]
  \centering
  \small
  \resizebox{\linewidth}{!}{%
  \def \dig {3}
  \npdecimalsign{.}
  \nprounddigits{\dig}
  \begin{tabular}{ln{1}{\dig}n{1}{\dig}|n{1}{\dig}n{1}{\dig}n{1}{\dig}}
      \toprule
      & \multicolumn{2}{c}{\textbf{Sequential}} & \multicolumn{3}{c}{\textbf{Global}}  \\ 
      & \textbf{ATE} & \textbf{HDG} & \textbf{ATE} & \textbf{HDG} & \textbf{MME} \\ 

      \midrule
      \textbf{ICP} 	        &  0.306785765141 & 0.013779635623932354  &  1.06416701722 & 0.055492743817969335 & 1.14863877\\

      \textbf{Noise}        &  0.214290821592 & 0.008396629320421780  &  0.362564829984 & 0.054228794783547696 & 1.14168757\\

      \textbf{Interp}       &  0.193716219841	& 0.022164274752721033  &  0.117693212713 & 0.0059785407935916794 & 0.22344477\\

      \textbf{Rotate}       &  0.186897625914	& 0.015127936720693408  &  0.120302920261 & 0.0047975743728047304 & 0.26751211\\

      \textbf{IMRP}         &  0.183418760279	& 0.011339709906020604  &  0.108129355885 &  0.0052473519970398074 & 0.19904358\\

      \textbf{Weight}       &  0.177255634923	& 0.011159007343393177  &  0.103527622905 & 0.0055023601050366933 & 0.27789450\\
      
      \textbf{Outliers}     &  0.162808820443 & 0.007183513034862473  &  0.0979268212538 & 0.00437102886078195 &  0.20279998\\
      \bottomrule
  \end{tabular}}
  \npnoround
  \caption{The absolute trajectory error (ATE), heading RMSE (HDG), mean map entropy (MME) measures after modifications, which are described in \refsection{sec:scan_matching}, are added to the standard ICP algorithm.
    The error metrics are calculated with respect to the ground truth trajectory from the tracking total station from the \textit{Chlumin dataset}.
    \label{tab:chlumin}
  }
  \figvspace
\end{table}


\subsection{Stara Voda church test flight}

The second indoor experimental verification was performed in the Stara Voda church.
We have chosen a larger church with different geometric layout and different sensor (Scanse Sweep) to verify the robustness of the proposed method.
Despite no available ground truth, the dataset features a loop closure (the same spot for takeoff and landing) that can be used to evaluate the error at the end of the trajectory.
This test flight illustrates the benefit of having a global position estimate that eliminates the drift by closing the loop at the end of the trajectory shown in \reffig{fig:chlumin_top}.
The Euclidean distance between the first and the last point of the trajectory is taken to describe the drift at the end of the trajectory.
While the drift of sequential matching was \SI{0.7925}{\metre}, the global matching improved the final error to \SI{0.3779}{\metre}.
The quality of the final map quantified by the MME measure is $0.3293$.
The higher MME score compared to the best values achieved in the \textit{Chlumin dataset} (\reftab{tab:chlumin}) is caused by the more noisy Scanse Sweep sensor and also the takeoff position was cluttered by a lot of people and small objects.



\subsection{Applications}
\label{sec:applications}

The proposed system was developed primarily for indoor navigation in churches and cathedrals where the insufficient lighting conditions prevent camera-based localization.
The historians are interested in photographs of murals and mosaics that are not reachable from the ground.
In addition to the photo documentation, the proposed light-weight solution of the localization based on 2D short-range LIDAR produces 3D models in the form of aligned point clouds from the mapping module.

One of the documentation missions was performed in Saint Anne Church in Stara Voda, which was damaged by Soviet soldiers during the occupation of the Czech Republic.
Several photographs of objects of interest were taken for the historians to assess the level of damage and state the necessary restoration works.
A virtual model of the church (shown in \reffig{fig:church_map}) was captured by the onboard LIDAR.

\begin{figure}
  \centering
      \begin{subfigure}{0.55\linewidth}
        \centering
        \adjincludegraphics[width=1.0\linewidth, trim={{0.0\width} {0.0\height} {0.0\width} {0.0\height}}, clip=true]{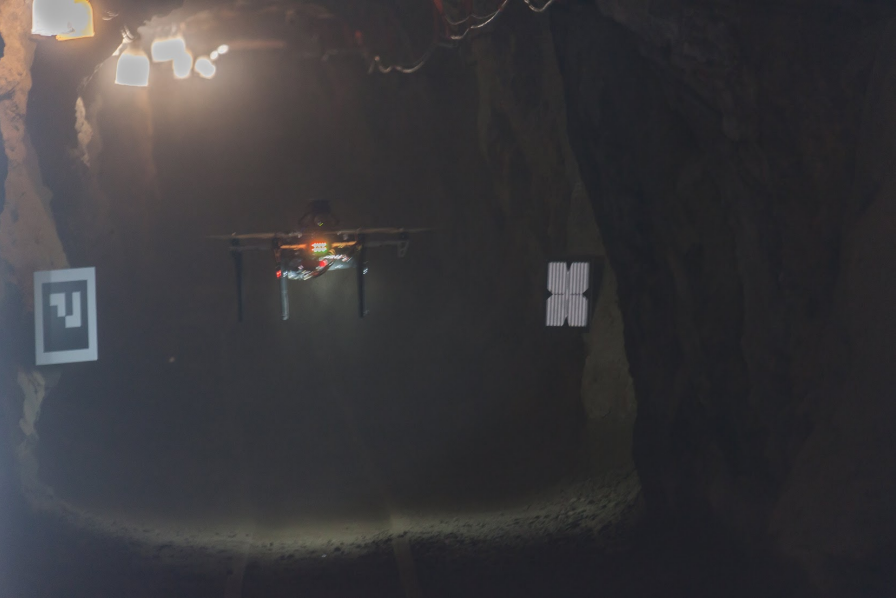}
      \end{subfigure}
      \begin{subfigure}{0.43\linewidth}
        \centering
        \adjincludegraphics[width=1.0\linewidth, trim={{0.0\width} {0.00\height} {0.0\width} {0.00\height}}, clip=true]{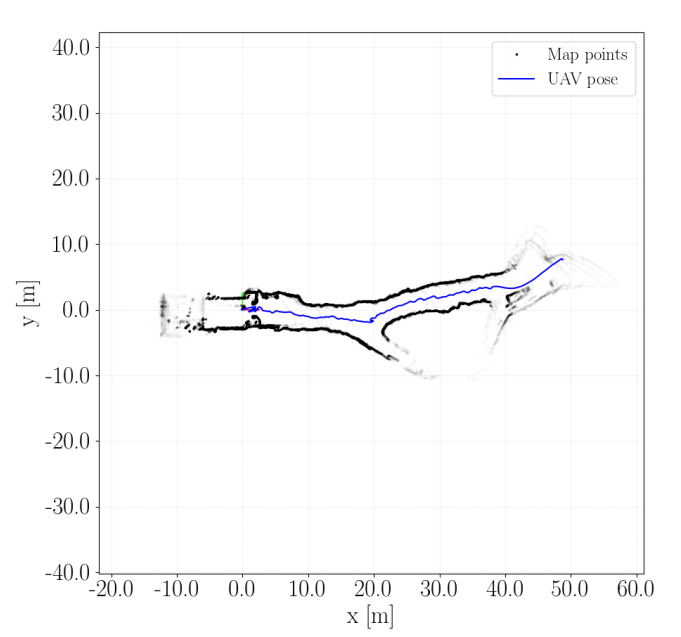}
      \end{subfigure}
  \caption{
    UAV equipped with a lightweight LIDAR navigates through a narrow mining tunnel with sensing degraded by dust (left). 
    A map of the tunnel is gradually built while simultaneously estimating the trajectory of the UAV (right).
    \label{fig:stix_map}
  }
  \figvspace
\end{figure}

Furthermore, the modularity of the system allowed easy adaptation and extension with navigation and path planning modules for deployment in DARPA Subterranean challenge.
The map in \reffig{fig:stix_map} was acquired during one of many flights through a mining tunnel at the DARPA Subterranean integration exercise.

\section{Conclusions}
\label{sec:conclusions}

We presented a pose estimation system for UAVs based on the alignment of laser scans from a lightweight 2D laser rangefinder as an alternative to large, heavy, and expensive 3D LIDARs.
The system is formed by a pipeline that fuses velocity estimate from sequential scan alignment with a global pose in a gradually built map to obtain a globally consistent smooth estimate, which is used by the controller to stabilize the UAV.

The proposed solution allows autonomous operation in indoor environments with insufficient illumination for camera-based solutions.
The large amount of applications where our UAV platforms were deployed in different kinds of environments, and with various sensor configurations were possible thanks to the proposed general and robust system.


\balance
\bibliographystyle{IEEEtran}
\bibliography{main}

\begin{thebibliography}{10}
\providecommand{\url}[1]{#1}
\csname url@samestyle\endcsname
\providecommand{\newblock}{\relax}
\providecommand{\bibinfo}[2]{#2}
\providecommand{\BIBentrySTDinterwordspacing}{\spaceskip=0pt\relax}
\providecommand{\BIBentryALTinterwordstretchfactor}{4}
\providecommand{\BIBentryALTinterwordspacing}{\spaceskip=\fontdimen2\font plus
\BIBentryALTinterwordstretchfactor\fontdimen3\font minus
  \fontdimen4\font\relax}
\providecommand{\BIBforeignlanguage}[2]{{%
\expandafter\ifx\csname l@#1\endcsname\relax
\typeout{** WARNING: IEEEtran.bst: No hyphenation pattern has been}%
\typeout{** loaded for the language `#1'. Using the pattern for}%
\typeout{** the default language instead.}%
\else
\language=\csname l@#1\endcsname
\fi
#2}}
\providecommand{\BIBdecl}{\relax}
\BIBdecl

\bibitem{lee2010geometric}
T.~Lee, M.~Leok, and N.~H. McClamroch, ``Geometric tracking control of a
  quadrotor uav on se(3),'' in \emph{CDC}, 2010, pp. 5420--5425.

\bibitem{mellinger2010minimum}
D.~Mellinger and V.~Kumar, ``Minimum snap trajectory generation and control for
  quadrotors,'' in \emph{ICRA}, 2011, pp. 2520--2525.

\bibitem{harik2016warehouse}
E.~H.~C. {Harik}, F.~{Guérin}, F.~{Guinand}, J.~{Brethé}, and
  H.~{Pelvillain}, ``Towards an autonomous warehouse inventory scheme,'' in
  \emph{SSCI}, 2016, pp. 1--8.

\bibitem{saska_etfa17}
M.~Saska, V.~Krátký, V.~Spurný, and T.~Báča, ``Documentation of dark areas
  of large historical buildings by a formation of unmanned aerial vehicles
  using model predictive control,'' in \emph{ETFA}, 2017, pp. 1--8.

\bibitem{cadena2016slam}
C.~{Cadena}, L.~{Carlone}, H.~{Carrillo}, Y.~{Latif}, D.~{Scaramuzza},
  J.~{Neira}, I.~{Reid}, and J.~J. {Leonard}, ``Past, present, and future of
  simultaneous localization and mapping: Toward the robust-perception age,''
  \emph{IEEE Transactions on Robotics}, vol.~32, no.~6, pp. 1309--1332, 2016.

\bibitem{delmerico2018vio}
J.~{Delmerico} and D.~{Scaramuzza}, ``A benchmark comparison of monocular
  visual-inertial odometry algorithms for flying robots,'' in \emph{ICRA},
  2018, pp. 2502--2509.

\bibitem{alismail2017lowlight}
H.~{Alismail}, M.~{Kaess}, B.~{Browning}, and S.~{Lucey}, ``Direct visual
  odometry in low light using binary descriptors,'' \emph{RAL}, vol.~2, no.~2,
  pp. 444--451, 2017.

\bibitem{horaud2016tof}
R.~Horaud, M.~Hansard, G.~Evangelidis, and C.~M{\'e}nier, ``An overview of
  depth cameras and range scanners based on time-of-flight technologies,''
  \emph{Machine Vision and Applications}, vol.~27, no.~7, pp. 1005--1020, 2016.

\bibitem{opromolla2016lidar}
R.~Opromolla, G.~Fasano, G.~Rufino, M.~Grassi, and A.~Savvaris,
  ``Lidar-inertial integration for uav localization and mapping in complex
  environments,'' in \emph{ICUAS}, 2016, pp. 649--656.

\bibitem{wang2013comprehensive}
F.~WANG, J.-Q. CUI, B.-M. CHEN, and T.~H. LEE, ``A comprehensive uav indoor
  navigation system based on vision optical flow and laser fastslam,''
  \emph{Acta Automatica Sinica}, vol.~39, no.~11, pp. 1889 -- 1899, 2013.

\bibitem{montemerlo2007fastslam}
M.~Montemerlo and S.~Thrun, \emph{FastSLAM: A Scalable Method for the
  Simultaneous Localization and Mapping Problem in Robotics}.\hskip 1em plus
  0.5em minus 0.4em\relax Springer-Verlag, 2007.

\bibitem{cui2014autonomous}
J.~Q. Cui, S.~Lai, X.~Dong, P.~Liu, B.~M. Chen, and T.~H. Lee, ``Autonomous
  navigation of uav in forest,'' in \emph{ICUAS}, 2014, pp. 726--733.

\bibitem{thrun2006graph}
S.~Thrun and M.~Montemerlo, ``The graph slam algorithm with applications to
  large-scale mapping of urban structures,'' \emph{The International Journal of
  Robotics Research}, vol.~25, no. 5-6, pp. 403--429, 2006.

\bibitem{lu97imrp}
F.~Lu and E.~Milios, ``Robot pose estimation in unknown environments by
  matching 2d range scans,'' \emph{Journal of Intelligent and Robotic Systems},
  vol.~18, no.~3, pp. 249--275, 1997.

\bibitem{bosse2012zebedee}
M.~Bosse, R.~Zlot, and P.~Flick, ``Zebedee: Design of a spring-mounted 3-d
  range sensor with application to mobile mapping,'' \emph{IEEE Transactions on
  Robotics}, vol.~28, no.~5, pp. 1104--1119, 2012.

\bibitem{razlaw2015evaluation}
J.~{Razlaw}, D.~{Droeschel}, D.~{Holz}, and S.~{Behnke}, ``Evaluation of
  registration methods for sparse 3d laser scans,'' in \emph{ECMR}, 2015, pp.
  1--7.

\bibitem{besl92icp}
P.~J. Besl and N.~D. McKay, ``A method for registration of 3-d shapes,''
  \emph{IEEE Transactions on Pattern Analysis and Machine Intelligence},
  vol.~14, no.~2, pp. 239--256, 1992.

\bibitem{segal2009gicp}
A.~Segal, D.~Hähnel, and S.~Thrun, ``Generalized-icp,'' 2009.

\bibitem{magnusson2007ndt}
M.~Magnusson, A.~Lilienthal, and T.~Duckett, ``Scan registration for autonomous
  mining vehicles using 3d-ndt,'' \emph{Journal of Field Robotics}, vol.~24,
  no.~10, pp. 803--827, 2007.

\bibitem{stoyanov2012ndt}
T.~Stoyanov, M.~Magnusson, H.~Andreasson, and A.~J. Lilienthal, ``Fast and
  accurate scan registration through minimization of the distance between
  compact 3d ndt representations,'' \emph{The International Journal of Robotics
  Research}, vol.~31, no.~12, pp. 1377--1393, 2012.

\bibitem{sturm12iros}
J.~Sturm, N.~Engelhard, F.~Endres, W.~Burgard, and D.~Cremers, ``A benchmark
  for the evaluation of rgb-d slam systems,'' in \emph{IROS}, 2012.

\bibitem{droeschel2014mme}
D.~{Droeschel}, J.~{Stückler}, and S.~{Behnke}, ``Local multi-resolution
  representation for 6d motion estimation and mapping with a continuously
  rotating 3d laser scanner,'' in \emph{ICRA}, 2014, pp. 5221--5226.

\bibitem{hornung13auro}
A.~Hornung, K.~M. Wurm, M.~Bennewitz, C.~Stachniss, and W.~Burgard,
  ``{OctoMap}: An efficient probabilistic {3D} mapping framework based on
  octrees,'' \emph{Autonomous Robots}, 2013.

\bibitem{zhang2014loam}
J.~Zhang and S.~Singh, ``Loam: Lidar odometry and mapping in real-time.'' in
  \emph{Robotics: Science and Systems}, vol.~2, no.~9, 2014, inproceedings.

\bibitem{shan2020lio}
T.~Shan, B.~Englot, D.~Meyers, W.~Wang, C.~Ratti, and D.~Rus, ``Lio-sam:
  Tightly-coupled lidar inertial odometry via smoothing and mapping,''
  \emph{arXiv preprint arXiv:2007.00258}, 2020.

\bibitem{jeff06frmsd}
J.~M. Phillips, R.~Liu, and C.~Tomasi, ``Outlier robust {ICP} for minimizing
  fractional {RMSD},'' \emph{CoRR}, vol. abs/cs/0606098, 2006.

\end{thebibliography}


\end{document}